\documentclass[a4paper,fleqn]{cas-dc}

\usepackage[numbers]{natbib}

\usepackage{booktabs}
\usepackage{amsmath}
\usepackage{placeins}
\usepackage{amssymb}
\usepackage{graphicx}
\usepackage[caption=false,font=footnotesize]{subfig}
\usepackage{xcolor}
\usepackage{makecell}
\usepackage{multirow}

\usepackage{enumitem}
\usepackage{siunitx}
\usepackage{xparse}
\usepackage{capt-of}
\usepackage{tabularx}

\usepackage{hyperref}
\hypersetup{
  colorlinks=true,
  allcolors=blue,
  linkcolor=blue,
  citecolor=blue,
  urlcolor=blue,
  filecolor=blue,
  breaklinks=true}
\AtBeginDocument{\hypersetup{
  colorlinks=true,
  allcolors=blue,
  linkcolor=blue,
  citecolor=blue,
  urlcolor=blue,
  filecolor=blue}}

\def\tsc#1{\csdef{#1}{\textsc{\lowercase{#1}}\xspace}}
\tsc{WGM}
\tsc{QE}

\begin{document}
\let\WriteBookmarks\relax
\def\floatpagepagefraction{1}
\def\textpagefraction{.001}
\shorttitle{M$^2$LG-DG for Cross-site MDD Classification}
\shortauthors{M.A. Hasan et~al.}

\title [mode = title]{M$^2$LG-DG: A Multi-modal Local-Global Domain Generalization Framework for Cross-site Major Depressive Disorder Classification}                      


\author[1]{Muhammad Asif Hasan}[orcid=0000-0003-3133-9106]
\credit{Writing -- original draft, Software, Investigation}
\ead{muhammadasif.hasan@griffithuni.edu.au}

\author[1]{Yanming Zhu}[orcid=0000-0002-8238-8090]
\credit{Validation, Supervision}
\ead{yanming.zhu@griffith.edu.au}

\author[1]{Xuefei Yin}[orcid=0000-0002-5784-7419]
\credit{Validation, Investigation, Supervision}
\ead{x.yin@griffith.edu.au}

\author[1]{Alan Wee-Chung Liew}[orcid=0000-0001-6718-7584]
\cormark[1]
\credit{Writing -- review \& editing, Project administration}
\ead{a.liew@griffith.edu.au}

\affiliation[1]{organization={School of Information and Communication Technology, Griffith University},
                city={Gold Coast},
                state={QLD},
                postcode={4215},
                country={Australia}}

\cortext[cor1]{Corresponding author}

\begin{abstract}
Classification models trained on resting-state functional magnetic resonance
imaging (rs-fMRI) often show reduced performance at imaging sites that were not observed during development, which limits their usefulness for clinical
deployment. Domain generalization (DG) aims to address this issue by learning
representations from source sites that remain effective for unseen target sites.
However, existing DG methods for psychiatric disorder classification commonly rely on a single imaging modality and often treat same-class subjects similarly when organizing the representation space. This overlooks the fact that subjects scanned at the same site already share scanner hardware, acquisition parameters, and preprocessing artifacts, so the learned representations become organized by acquisition condition rather than by diagnostic label. In this work, we present M$^{2}$LG-DG, a source-only multi-modal local-global framework for cross-site major depressive disorder (MDD) classification. The framework uses a dual-stream rs-fMRI encoder, in which the global pathway models inter-regional dependencies through self-attention and the local pathway performs graph-constrained aggregation over functional connectivity-derived brain graphs. Imaging and non-imaging representations are decomposed into shared and private components and integrated through bidirectional cross-attention with a learned modality gate. In addition, a cross-site supervised contrastive objective defines positive pairs using same-class subjects from different source sites, encouraging the fused representation to capture diagnostic information across acquisition domains. On four held-out REST-meta-MDD sites, M$^{2}$LG-DG achieves 69.48\% AUC and exceeds the closest comparison method by 2.18 percentage points. Additional experiments on the Autism Brain Imaging Data Exchange (ABIDE) dataset support its
applicability to other psychiatric neuroimaging tasks.
\end{abstract}





\maketitle

\section{Introduction}
\label{sec:Intro}

Major depressive disorder (MDD) is one of the most prevalent neuropsychiatric conditions, affecting approximately 185 million individuals worldwide, with an estimated lifetime prevalence of 10\%~\cite{marx2023major}. Early and accurate identification of MDD is important, because psychiatric disorders contribute substantially to the global disease burden~\cite{gibbs2018toward}. Resting-state functional magnetic resonance imaging (rs-fMRI) provides a non-invasive way to characterize functional brain organization by measuring blood oxygen level-dependent (BOLD) signal fluctuations across brain regions. Functional connectivity (FC), which is commonly estimated from correlations among regional BOLD time series, has been widely used to study neuropsychiatric disorders, because altered connectivity patterns are associated with clinical manifestations and may serve as candidate biomarkers~\cite{preti2017dynamic,li2025interpretable}. Although rs-fMRI based computer-aided classification has shown promising results in controlled settings~\cite{kong2024multi, liu2023deep, liu2025dynamic}, its practical use is limited by poor generalization across imaging sites. Models trained and tested within the same site can perform well~\cite{wang2026adhd,lee2024}, but scanner hardware, acquisition protocols, repetition time, scan duration, preprocessing choices, recruitment criteria, and participant demographics vary substantially across centers. These variations can lead a model to rely on site-specific characteristics rather than disease-related functional patterns, which results in unreliable predictions when the model is applied to an unseen site~\cite{fang2023unsupervised,shi2025novel}.

Domain adaptation (DA) and domain generalization (DG) are two common strategies for addressing cross-site distribution shift in multi-site neuroimaging studies~\cite{su2024m2dc}. DA methods reduce the discrepancy between source and target domains by using target-domain data during training, which is less suitable for deployment settings in which a model must be applied to a new clinical site that was not observed during model development. DG offers a more realistic alternative by learning representations from source domains only, provided that the target domain remains entirely unseen during training, validation, and model selection. Existing DG methods for medical and neuroimaging applications have explored domain-invariant representation learning~\cite{liu2023d}, feature disentanglement~\cite{gholami2023latent}, data augmentation~\cite{lehner20223d}, and meta-learning~\cite{li2018learning}. Most DG studies for psychiatric disorder classification, however, rely mainly on a single imaging modality, typically rs-fMRI~\cite{su2024m2dc,qiu2025metaexplainer}, and therefore make limited use of subject-level information such as age, sex, and years of education. Prior studies have shown that combining imaging and non-imaging data can improve the characterization of psychiatric disorders~\cite{cai2025mm}, and recent work on multi-modal DG further supports the value of integrating heterogeneous information sources~\cite{dong2026advances}. Non-imaging variables must nevertheless be handled with care in multi-site psychiatric datasets, because they may reflect not only clinically relevant variation but also site-specific recruitment patterns and cohort-level differences. An effective multi-modal DG framework should therefore integrate imaging and non-imaging information in a structured manner, so that complementary diagnostic information is preserved while the influence of site-related correlations is reduced.

Incorporating rs-fMRI and non-imaging data into a unified framework raises three challenges. The \textbf{\textit{first issue}} is the integration of modalities that contain both shared and modality-specific information~\cite{dong2024towards}. rs-fMRI captures spatiotemporal brain activity and connectivity alterations, whereas non-imaging variables describe subject-level factors that are not directly measured by brain imaging. Mapping both modalities into a single common space may retain shared information, but it may also suppress modality-specific cues that are useful for classification. The \textbf{\textit{second issue}} arises within the rs-fMRI modality itself. In this work, global modeling refers to unrestricted interaction among region-level temporal representations, and local modeling refers to graph-constrained aggregation over FC-derived neighborhoods. The two views are complementary, since global modeling captures distributed dependencies across distant regions, whereas local modeling focuses on connectivity patterns within functionally related neighborhoods~\cite{liu2025dynamic}. Most existing methods emphasize either temporal dynamics or graph-based spatial relationships, but joint modeling of
local and global characteristics may provide a more complete representation of disease-related brain alterations. The \textbf{\textit{third issue}} concerns how the learned representations are organized across sites. Subjects scanned at the same
site share acquisition conditions such as scanner hardware, field strength, and preprocessing artifacts, and these shared conditions can cause representations to cluster by site rather than by diagnostic label. Contrastive learning is a natural way to shape the representation space~\cite{gomariz2025joint}, but standard supervised contrastive learning treats every same-class subject as a positive, including same-class subjects from the same site. Such pairs are already similar because of shared scanner characteristics rather than disease-related patterns, so treating them as positives reinforces site-dependent structure and works against cross-site generalization.

To address these challenges, we propose a multi-modal local-global domain generalization framework (M$^2$LG-DG) for cross-site MDD classification using
rs-fMRI and non-imaging data. First, to integrate modalities that carry both shared and modality-specific information, we develop a shared-private multi-modal fusion module. The imaging and non-imaging embeddings, obtained respectively from a dual-stream rs-fMRI encoder and a multilayer perceptron (MLP), are decomposed into a shared component that captures cross-modal agreement and a private component that preserves cues available in only one modality. The decomposed representations are then combined through bidirectional cross-attention with a learned modality gate, which allows each modality to attend to complementary evidence in the other while the gate regulates their relative contribution for each subject. Second, to jointly characterize local and global functional patterns, we design a dual-stream rs-fMRI encoder with two pathways that process spatial and temporal information in complementary orders. The global pathway first extracts per-region-of-interest (ROI) temporal features using one-dimensional convolutions and then models cross-region dependencies using Transformer-based self-attention, so that long-range interactions are learned without a predefined graph constraint. The local pathway follows the opposite order, and first aggregates spatial information over an FC-derived brain graph using Chebyshev graph convolutions and then models temporal evolution through temporal convolutions, which emphasizes coherent patterns within functionally related neighborhoods. Third, to prevent the learned representations from being organized by acquisition condition rather than by diagnostic label, we formulate a cross-site supervised contrastive objective on the fused representation. For each anchor subject, positive pairs are formed only from same-class subjects acquired at different source sites, and same-site same-class subjects are excluded from the positive set. This restriction encourages the model to align representations according to diagnostic label across sites, which supports generalization to a target site that is unseen during training.

The main contributions of this work are as follows:
\begin{enumerate}[label={\arabic*)}]
    \item We propose M$^2$LG-DG, a source-only multi-modal DG framework for cross-site MDD classification that integrates rs-fMRI and non-imaging variables while keeping each target site entirely unseen during training, validation, and hyperparameter selection.

    \item We design a dual-stream rs-fMRI encoder that captures complementary local and global functional patterns. The global pathway models unrestricted interactions among ROI-level temporal representations, while the local pathway performs graph-constrained spatiotemporal modeling over FC-derived brain networks.

    \item We construct a structured multi-modal fusion strategy based on shared-private decomposition, bidirectional cross-attention, and a learned modality gate, which separates cross-modal and modality-specific information
    rather than relying on feature concatenation.

    \item We formulate a cross-site supervised contrastive objective that excludes same-site same-class pairs from the positive set, so that the fused representation is organized by diagnostic label rather than by site-specific acquisition conditions. Experiments on the REST-meta-MDD dataset under a held-out-site protocol, together with additional experiments on the ABIDE dataset, demonstrate the effectiveness of the proposed framework.
\end{enumerate}

\section{Related work}
\label{sec:R_W}

\subsection{Domain generalization}

DG aims to learn representations from multiple source domains that remain effective when applied to previously unseen target domains~\cite{dong2024towards}. Existing DG methods can be broadly organized into three categories: data manipulation, representation learning, and learning strategy-based approaches~\cite{dong2026advances}.

Data manipulation methods alter the training data to encourage the learning of more transferable representations. This is typically achieved through data augmentation~\cite{ouyang2022causality} or data generation~\cite{na2025bridging}, both of which increase the diversity of the training distribution and can improve generalization. Representation learning methods, on the other hand, aim to identify feature spaces that are stable across domains, enabling the model to retain task-relevant information when transferred to new environments. For instance, \cite{zhang2023deep} proposed an information bottleneck-based DG framework that encourages domain-invariant representations through a combination of maximum feature entropy, class-wise instance discrimination, inter-dimension decorrelation, and intra-dimension uniformity. Other approaches have sought to reduce distributional variation across source domains through statistical alignment objectives such as maximum mean discrepancy~\cite{fang2023unsupervised} and CORAL loss~\cite{sun2016deep}, while batch normalization-based strategies have also been explored to obtain representations that are less sensitive to domain-specific statistics~\cite{nam2018batch}. In a related direction, \cite{arjovsky2019invariant} introduced invariant risk minimization, which promotes representations for which the optimal classifier remains consistent across all training domains. Learning strategy-based methods constitute the third category and focus on the optimization procedure, training schedule, or domain-sampling strategy used to improve generalization. \cite{su2024m2dc} proposed M$^{2}$DC, which integrates meta batch normalization with a distance constraint under a meta-learning framework to learn more discriminative brain graph representations for cross-site MDD classification. \cite{qiu2025metaexplainer} proposed MetaExplainer, a method that identifies domain-agnostic connectome alterations to support both accurate and interpretable neuropsychiatric disorder classification across clinical centers.

In the context of psychiatric disorder classification, several studies have adopted DA to mitigate cross-site distribution differences~\cite{ren2026unsupervised,ma2026augmentation}. However, DA methods require access to target-domain data during training, which restricts their applicability when data from new clinical sites are not available at the time of model development. DG methods are therefore more suitable for deployment scenarios where models must generalize to previously unseen acquisition environments. In this work, we adopt the DG setting, where the model is trained exclusively on source-domain data and must generalize to entirely unseen target sites without access to their data during training, validation, or hyperparameter selection.

\subsection{Multi-modal representation learning}

Multi-modal representation learning aims to learn informative feature representations from two or more modalities for downstream tasks~\cite{liu2025mtif}. This direction has led to the development of multi-modal domain generalization (M$^2$DG), which seeks to improve generalization to unseen domains by integrating complementary information from multiple modalities~\cite{dong2026advances}. Compared with unimodal settings, M$^2$DG introduces additional challenges. First, different modalities may experience heterogeneous domain shifts, meaning that each modality can be affected by different types and degrees of distribution change. Second, preserving cross-modal dependencies during alignment is difficult, since aligning each modality independently may weaken the relationships between modalities. Third, multi-modal settings often involve practical issues such as modality imbalance and missing modalities. These characteristics suggest that M$^2$DG should be considered distinct from unimodal DG.

Existing multi-modal learning methods generally follow two main directions. Approaches such as CLIP~\cite{wasim2023vita} and ALIGN~\cite{he2023align} use separate encoders for different modalities and employ contrastive learning to map their features into a common embedding space. In contrast, unified models~\cite{wang2022ofa,bao2022vlmo} convert different modalities into token sequences~\cite{bao2021beit} and apply a single Transformer for joint representation learning. Although these methods have shown strong performance, they primarily emphasize information shared across modalities and may not fully preserve modality-specific characteristics. In psychiatric disorder diagnosis, this issue is particularly relevant because each modality carries both common diagnostic information and distinct complementary cues. For example, rs-fMRI captures spatiotemporal patterns of neural activity and connectivity alterations across brain regions, whereas non-imaging variables reflect clinical and physiological risk factors that are not directly observable from brain imaging alone. Projecting both modalities into a single shared space may retain diagnostic signals common to both modalities, but it can also reduce modality-specific details contributed by each source. To address this limitation, we decompose the features of each modality into shared and modality-specific components, allowing common information to be aligned across modalities while preserving distinctive information from each source.

\subsection{Contrastive learning for domain-invariant representation}

Contrastive learning has become an effective approach for learning discriminative feature representations and has been applied in several studies to address multi-site heterogeneity in medical imaging. Barbano et al. \cite{barbano2023contrastive} proposed a contrastive regression loss for brain age prediction from multi-site MRI data and demonstrated improved robustness to site-related noise compared with standard supervised approaches. Wang et al. \cite{wang2022contrastive} formulated a contrastive functional connectivity graph learning method for population-based fMRI classification, in which representations of same-class subjects are attracted while those of different-class subjects are repelled. Zhi et al. \cite{zhi2024supervised} proposed a supervised contrastive learning-based DG network for cross-subject motor decoding, where supervised contrastive learning was combined with domain-agnostic mixup to achieve class-level alignment across subjects. In a related direction, \cite{xu2022adversarial} proposed an adversarial domain synthesizer with mutual information regularization to learn semantically consistent synthetic domains for single-source organ segmentation under unseen modalities, protocols, and scanner sites.

Despite these advances, most existing contrastive learning methods do not explicitly account for the relationship between class labels and site membership when constructing positive and negative pairs. In multi-site brain imaging studies, two subjects from the same diagnostic class and the same site may appear similar in the learned representation because they share scanner-specific characteristics, such as field strength, acquisition parameters, or preprocessing artifacts, rather than disease-related patterns. Including such pairs as positives may encourage the model to encode site-related information instead of disease-relevant features. To address this limitation, we propose a cross-site variant of supervised contrastive learning that excludes same-site same-class pairs from the positive set. The detailed formulation is presented in Section~\ref{sec:classification_contrastive}.

\begin{figure*}[htbp] 
\centerline{\includegraphics[width=1.0\textwidth]{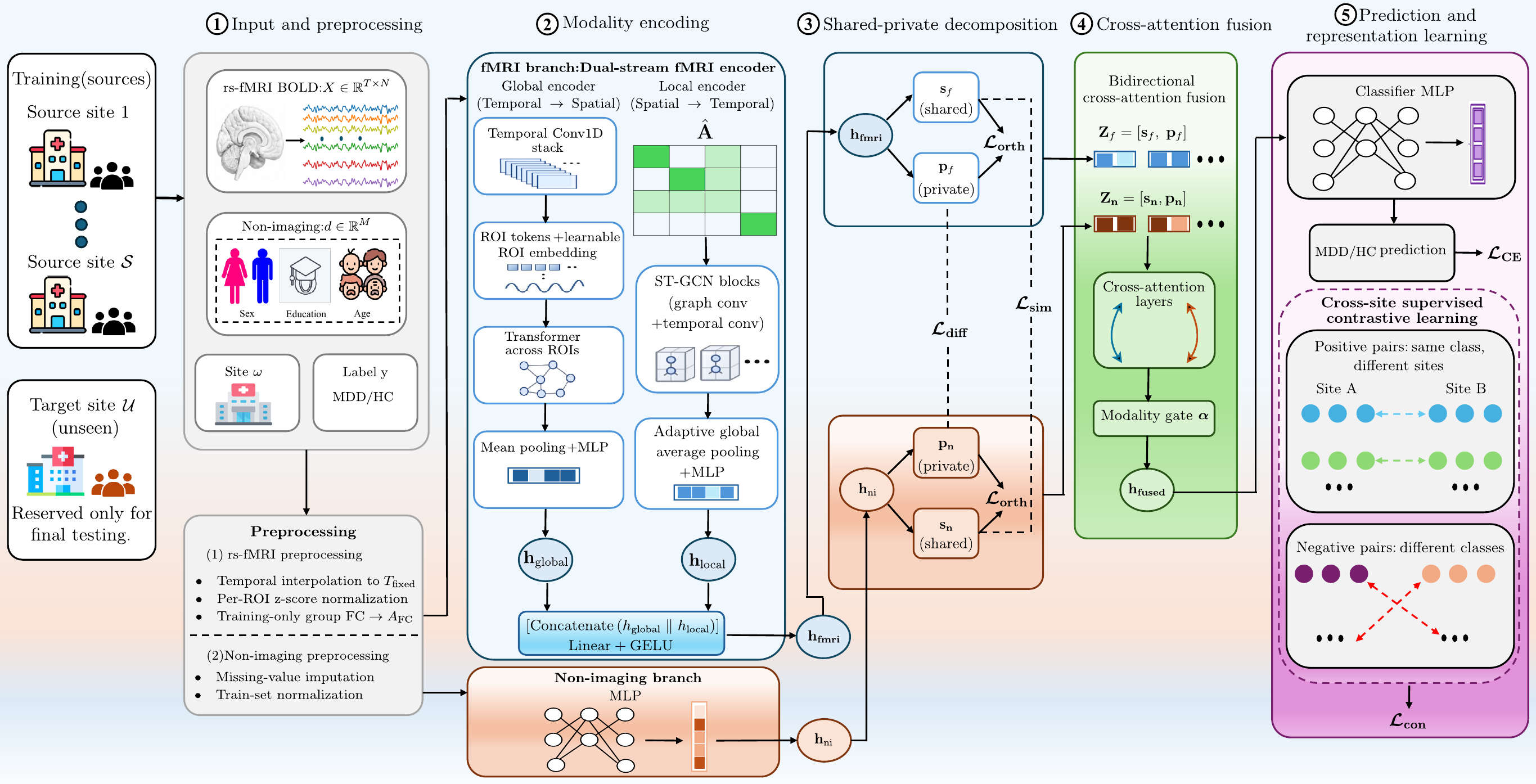}}
\caption{Overview of the proposed M$^2$LG-DG framework. In Stage~1, multi-site rs-fMRI time series and non-imaging variables are preprocessed and provided to the dual-stream fMRI encoder in Stage~2, which includes a global encoder (temporal $\rightarrow$ spatial using 1D CNN and Transformer layers) and a local encoder (spatial $\rightarrow$ temporal using Chebyshev graph convolutions), together with a non-imaging MLP encoder. In Stage~3, the learned representations are separated into shared and private components and guided by similarity, orthogonality, and difference losses. In Stage~4, the decomposed tokens are integrated using bidirectional cross-attention and a learned modality gate that assigns sample-specific weights to each modality. Finally, in Stage~5, the fused representation $h_{\mathrm{fused}}$ is employed for MDD/HC classification through $\mathcal{L}_{\mathrm{CE}}$ and cross-site supervised contrastive learning $\mathcal{L}_{\mathrm{con}}$, where same-site same-class pairs are excluded from the positive set to support site-invariant disease representation learning.}
\label{fig1}
\end{figure*} 

\section{Proposed method}
\label{sec:Proposed_Method}

\subsection{Problem definition and overall architecture}
\label{sec:problem_definition}

The overall framework of Multi-modal Local-Global Domain Generalization (M$^2$LG-DG) is shown in Fig.~\ref{fig1}. The framework consists of five main stages. First, multi-site rs-fMRI time series and non-imaging variables are preprocessed. Second, the preprocessed rs-fMRI data are encoded by a dual-stream fMRI encoder, while the non-imaging variables are encoded by a non-imaging MLP encoder. The dual-stream fMRI encoder includes a global pathway that follows a temporal-to-spatial order using one-dimensional convolutional neural network (1D CNN) layers and Transformer layers, and a local pathway that follows a spatial-to-temporal order using Chebyshev graph convolutions. Third, the learned imaging and non-imaging representations are decomposed into shared and private components and regularized using similarity, orthogonality, and difference losses. Fourth, the decomposed tokens are integrated through bidirectional cross-attention and a learned modality gate, which assigns sample-specific weights to each modality. Finally, the fused representation $\mathbf{h}_{\mathrm{fused}}$ is used for MDD versus healthy control (HC) classification through the classification loss $\mathcal{L}_{\mathrm{CE}}$ and the cross-site supervised contrastive loss $\mathcal{L}_{\mathrm{con}}$.

For each subject, the brain is parcellated into $N$ ROIs using a predefined atlas, and the average BOLD time course is extracted from each region. The resulting fMRI data are denoted as $\mathbf{X} \in \mathbb{R}^{T \times N}$, where $T$ is the number of time points and $N$ is the number of ROIs. Each subject is also associated with a non-imaging feature vector $\mathbf{d} \in \mathbb{R}^{M}$, where $M$ denotes the number of non-imaging variables, such as sex, age, and years of education. We consider a DG setting with $S$ source domains and one unseen target domain. The source data are defined as
\begin{equation}
    \mathcal{S}
    =
    \bigl\{
    (\mathbf{X}_i,\mathbf{d}_i,y_i,\omega_i)
    \bigr\}_{i=1}^{N_S},
    \label{eq:dg_setting}
\end{equation}
where $y_i \in \{0,1\}$ is the diagnostic label, $\omega_i \in \{1,\ldots,S\}$ is the source-domain index, and $N_S$ denotes the total number of source samples. The condition $\mathcal{U} \cap \mathcal{S} = \emptyset$ indicates that the unseen target domain $\mathcal{U}$ is disjoint from the source domains. The proposed framework aims to learn diagnosis-related representations that generalize to samples from unseen target domains. 

\subsection{Data preprocessing}
\label{sec:data_preprocessing}

\subsubsection{rs-fMRI temporal standardization}
\label{sec:fmri_temporal_standardization}
The preprocessed rs-fMRI time series obtained from different sites contained varying numbers of time points because of differences in scan duration and repetition time (TR). Prior to release, all time series had been band-pass filtered to $0.01$--$0.1$~Hz at the participating sites, as described in Section~\ref{sec:Data_MDD}. Since the proposed encoders require a fixed input length, each ROI time series was independently resampled to $T_{\mathrm{fixed}}$ samples using cubic spline interpolation, with $T_{\mathrm{fixed}} = 200$ used in all experiments. This resampling provides a consistent input tensor dimension across subjects. The procedure was also performed independently for each subject without using information from other subjects, diagnostic labels, or the held-out target domain. The sensitivity of the proposed framework to $T_{\mathrm{fixed}}$ is examined in Section~\ref{sec:sensitivity_study}. Following resampling, each ROI time series was normalized within each subject using z-score normalization:
\begin{equation}
    \tilde{x}_{i,m,n} =
    \frac{x_{i,m,n} - \mu_{i,n}}{\sigma_{i,n} + \epsilon},
    \label{eq:roi_zscore}
\end{equation}
where $x_{i,m,n}$ denotes the resampled signal of ROI $n$ for subject $i$ at
grid position $m$, $\mu_{i,n}$ and $\sigma_{i,n}$ denote the corresponding mean
and standard deviation, respectively, and $\epsilon$ is a small constant added
for numerical stability.

\subsubsection{Non-imaging data preprocessing}
\label{sec:nonimaging_preprocessing}
The non-imaging metadata contained a small number of invalid entries.
To prevent information leakage from the held-out target domain, all
imputation and normalization parameters were estimated exclusively from
the source-training subjects within each held-out-site fold and without
using diagnostic labels. Invalid sex values were replaced by the mode
estimated from the source-training subjects. For subjects belonging to
source sites, invalid age values were replaced by the median age of valid
source-training subjects from the corresponding site; when no valid age
was available for that site, the overall median of the source-training
subjects was used. Invalid age values in the held-out target site were
imputed using only the overall source-training median. Zero-valued years
of education were retained as valid observations. After imputation, the
non-imaging features were standardized using the mean and standard
deviation estimated from the source-training subjects, and the same
parameters were subsequently applied unchanged to the source-validation
and target subjects.

\subsubsection{Functional connectivity graph construction}
\label{sec:fc_graph_construction}
The brain graph used by the local encoder was constructed from FC estimated exclusively from the non-target source
subjects. For each held-out target site, Pearson correlations were computed
between every pair of ROI time series for each source subject. The resulting
subject-level FC matrices were first averaged across all source subjects,
after which the magnitude of the group-level FC matrix was thresholded to
construct a binary adjacency matrix:
\begin{equation}
\begin{aligned}
    \mathbf{F}
    &=
    \frac{1}{N_{\mathrm{src}}}
    \sum_{i=1}^{N_{\mathrm{src}}}
    \mathrm{corr}(\tilde{\mathbf{X}}_i),
    \\
    \mathbf{A}_{\mathrm{FC}}
    &=
    \mathbb{1}
    \left[
        |\mathbf{F}|
        \geq
        \tau_{\rho}
    \right]
    \vee
    \mathbf{I},
\end{aligned}
\label{eq:group_fc}
\end{equation}
where $N_{\mathrm{src}}$ denotes the number of source subjects,
$\mathrm{corr}(\cdot)$ returns the pairwise ROI correlation matrix, and
$\mathbb{1}[\cdot]$ denotes the elementwise indicator function. The
diagonal entries of $\mathbf{F}$ were set to zero before determining the
threshold. The scalar $\tau_{\rho}$ denotes the $\rho$-th percentile of
the nonzero off-diagonal entries of $|\mathbf{F}|$, such that only the
strongest group-level connections are retained.

Because the absolute value is applied after averaging the subject-level
FC matrices, the resulting graph preferentially retains connections that
are strong and directionally consistent across source subjects, whereas
correlations with opposing signs across subjects may partially cancel
before thresholding. Finally, self-loops are introduced through the
elementwise disjunction with the identity matrix $\mathbf{I}$, yielding
the binary FC-derived adjacency matrix
$\mathbf{A}_{\mathrm{FC}}\in\{0,1\}^{N\times N}$.

\subsection{Dual-stream fMRI encoder}
\label{sec:dual_stream_encoder}

The rs-fMRI data contain temporal information that describes regional activity changes over time and spatial information that describes interactions among brain regions. To represent both aspects, we adopt a dual-stream encoding strategy in which two encoders process the fMRI data using different orders of temporal and spatial operations. The global encoder first extracts temporal features and then models relationships among ROIs. The local encoder first aggregates graph-based spatial information and then models temporal evolution. The outputs of the two encoders are combined to form a single fMRI representation.

\subsubsection{Global encoder}
\label{sec:global_encoder}

The global encoder processes the temporal dimension first and then models relationships across ROIs. Given the input $\tilde{\mathbf{X}} \in \mathbb{R}^{T_{\mathrm{fixed}} \times N}$, each ROI time series is treated as a one-dimensional signal and passed through a shared stack of temporal convolutions. We apply $L_c$ layers of one-dimensional convolutions with progressively increasing channel sizes $(C_1,C_2,\ldots,C_{L_c})$. Each convolutional layer is followed by batch normalization, Gaussian error linear unit (GELU) activation, max pooling with a factor of 2, and dropout. After the convolutional stack, the temporal dimension is reduced, and the output for each ROI is flattened and projected to a $d$-dimensional vector. The global encoding process is formulated as
\begin{equation}
\begin{aligned}
    &\mathbf{H}_{\mathrm{roi}}
    =
    \mathrm{Proj}
    \left(
        \mathrm{CNN}_{1D}(\tilde{\mathbf{X}})
    \right),
    \hspace{0.8em}
    \mathbf{Z}_{g}
    =
    \mathbf{H}_{\mathrm{roi}}+\mathbf{P},
    \\[2pt]
    &\mathbf{h}_{\mathrm{global}}
    =
    \mathrm{MLP}_{g}
    \left(
        \frac{1}{N}
        \sum_{n=1}^{N}
        \mathrm{Transformer}(\mathbf{Z}_{g})_{n}
    \right).
\end{aligned}
\label{eq:global_encoder}
\end{equation}
where $\mathbf{H}_{\mathrm{roi}} \in \mathbb{R}^{N \times d}$ is the ROI-level feature matrix and $\mathbf{P} \in \mathbb{R}^{N \times d}$ is a learnable ROI identity embedding added to preserve ROI identity. The Transformer encoder enables each ROI token to attend to all other ROI tokens without a predefined graph constraint. The average over the $N$ ROI tokens is passed through $\mathrm{MLP}_{g}$ to produce the global embedding $\mathbf{h}_{\mathrm{global}} \in \mathbb{R}^{D_g}$.

\subsubsection{Local encoder}
\label{sec:local_encoder}

The local encoder follows the opposite processing order. It first operates in the spatial domain using graph convolutions and then applies temporal convolutions. This pathway is implemented using a spatio-temporal graph convolutional network (ST-GCN). Let $\mathbf{A}_{\mathrm{FC}}$ denote the FC-derived adjacency matrix constructed from the source-training subjects. To allow the graph structure to be adjusted during training, the local encoder maintains learnable edge embeddings $\mathbf{E} \in \mathbb{R}^{N \times d_e}$. The learned adjacency matrix, the combined adjacency matrix, and the normalized adjacency operator are defined as
\begin{equation}
\begin{aligned}
    \mathbf{A}_{\mathrm{learn}}
    &=
    \mathrm{softmax}
    \!\left(
    \mathbf{E}\mathbf{E}^{\top}
    \right),
    \\[2pt]
    \mathbf{A}_{\mathrm{final}}
    &=
    \beta \mathbf{A}_{\mathrm{learn}}
    +
    (1-\beta)\mathbf{A}_{\mathrm{FC}},
    \\[2pt]
    \hat{\mathbf{A}}
    &=
    \mathbf{D}^{-1/2}
    \left(
    \mathbf{A}_{\mathrm{final}}+\mathbf{I}
    \right)
    \mathbf{D}^{-1/2},
\end{aligned}
    \label{eq:adjacency}
\end{equation}
where the softmax is applied row-wise, $\beta$ is a fixed mixing coefficient set to $0.5$, and $\mathbf{D}$ is the degree matrix of $\mathbf{A}_{\mathrm{final}}+\mathbf{I}$. The learned adjacency matrix provides a data-driven refinement of the connectivity pattern, while the FC-derived adjacency matrix supplies the group-level functional structure estimated from the source-training subjects. The symmetric normalization converts the aggregation into a degree-weighted average rather than a sum.

Spatial aggregation is performed by expanding the normalized adjacency operator in a Chebyshev polynomial basis of order $K$, following the formulation of~\cite{defferrard2016convolutional}. Let $\mathbf{H} \in \mathbb{R}^{N \times C_{\mathrm{in}}}$ denote the node feature matrix at a given timepoint of a given block. The graph convolution is computed as
\begin{equation}
\begin{aligned}
    \mathcal{T}_0(\hat{\mathbf{A}})\mathbf{H}
    &=
    \mathbf{H},
    \qquad
    \mathcal{T}_1(\hat{\mathbf{A}})\mathbf{H}
    =
    \hat{\mathbf{A}}\mathbf{H},
    \\[2pt]
    \mathcal{T}_k(\hat{\mathbf{A}})\mathbf{H}
    &=
    2\hat{\mathbf{A}}\,
    \mathcal{T}_{k-1}(\hat{\mathbf{A}})\mathbf{H}
    -
    \mathcal{T}_{k-2}(\hat{\mathbf{A}})\mathbf{H},
    \\[2pt]
    \mathbf{Y}
    &=
    \sum_{k=0}^{K-1}
    \mathcal{T}_k(\hat{\mathbf{A}})
    \mathbf{H}
    \mathbf{W}_k
    +
    \mathbf{b},
\end{aligned}
    \label{eq:chebyshev}
\end{equation}
where $\mathcal{T}_k(\cdot)$ denotes the $k$-th order Chebyshev polynomial evaluated at the normalized adjacency operator, $\mathbf{W}_k \in \mathbb{R}^{C_{\mathrm{in}} \times C_{\mathrm{out}}}$ are learnable weight matrices, and $\mathbf{b}$ is a bias term. The order $K$ determines the number of graph-neighborhood hops used for aggregation. The graph convolution is applied independently at each timepoint with shared weights. After spatial aggregation, temporal convolution is applied along the time dimension to model changes in graph-based spatial patterns. Each ST-GCN block consists of graph convolution, temporal convolution with batch normalization, addition of a residual branch, rectified linear unit activation, and dropout.

The local encoder consists of $L_s$ ST-GCN blocks with progressively increasing channel sizes and a temporal stride applied after the first block for downsampling. After the final block, adaptive global average pooling over the temporal and spatial dimensions produces a single vector, which is projected through an MLP:
\begin{equation}
    \mathbf{h}_{\mathrm{local}}
    =
    \mathrm{MLP}_{l}
    \left(
    \mathrm{Pool}(\mathbf{Y}^{(L_s)})
    \right),
    \label{eq:local_embedding}
\end{equation}
where $\mathbf{Y}^{(L_s)}$ is the output of the final ST-GCN block, $\mathrm{Pool}(\cdot)$ denotes adaptive global average pooling, and $\mathbf{h}_{\mathrm{local}} \in \mathbb{R}^{D_l}$ is the local embedding.

\subsubsection{fMRI representation fusion}
\label{sec:fmri_representation_fusion}

The global and local embeddings provide complementary fMRI representations. The global encoder models unrestricted inter-ROI dependencies, whereas the local encoder models graph-constrained functional neighborhoods. The two embeddings are concatenated and projected through a linear layer with GELU activation:
\begin{equation}
    \mathbf{h}_{\mathrm{fmri}}
    =
    \phi
    \left(
    \mathbf{W}_p
    [
    \mathbf{h}_{\mathrm{global}}
    \,\|\,
    \mathbf{h}_{\mathrm{local}}
    ]
    +
    \mathbf{b}_p
    \right),
    \label{eq:fmri_combine}
\end{equation}
where $\phi$ denotes the GELU activation, $\|$ denotes concatenation, and $\mathbf{W}_p$ and $\mathbf{b}_p$ are projection parameters. The resulting $\mathbf{h}_{\mathrm{fmri}} \in \mathbb{R}^{D_f}$ is used as the fMRI representation for multi-modal fusion.

\subsection{Non-imaging encoder}
\label{sec:non_imaging_encoder}

The preprocessed non-imaging features are encoded using an MLP with two hidden layers. Each hidden layer is followed by batch normalization, GELU activation, and dropout, and a final linear layer produces the output embedding. The non-imaging representation is computed as
\begin{equation}
    \mathbf{h}_{\mathrm{ni}}
    =
    \mathrm{MLP}_{\mathrm{ni}}
    (
    \tilde{\mathbf{d}}
    ),
    \label{eq:ni_encoder}
\end{equation}
where $\tilde{\mathbf{d}}$ denotes the preprocessed non-imaging feature vector, and $\mathbf{h}_{\mathrm{ni}} \in \mathbb{R}^{D_n}$ denotes the non-imaging embedding.

\subsection{Shared-private multi-modal decomposition}
\label{sec:shared_private_decomposition}

The fMRI and non-imaging representations may contain both shared information and modality-specific information. To preserve both forms of information, each modality representation is decomposed into a shared component and a private component. The shared components are encouraged to capture information common to both modalities, while the private components are encouraged to preserve modality-specific information. The decomposition is implemented using four separate two-layer MLPs:
\begin{equation}
\begin{aligned}
    \mathbf{s}_f
    &=
    \psi_f^{\mathrm{shared}}
    (
    \mathbf{h}_{\mathrm{fmri}}
    ),
    &
    \mathbf{p}_f
    &=
    \psi_f^{\mathrm{private}}
    (
    \mathbf{h}_{\mathrm{fmri}}
    ),
    \\[2pt]
    \mathbf{s}_n
    &=
    \psi_n^{\mathrm{shared}}
    (
    \mathbf{h}_{\mathrm{ni}}
    ),
    &
    \mathbf{p}_n
    &=
    \psi_n^{\mathrm{private}}
    (
    \mathbf{h}_{\mathrm{ni}}
    ),
\end{aligned}
    \label{eq:decomposition}
\end{equation}
where $\mathbf{s}_f,\mathbf{s}_n \in \mathbb{R}^{D_s}$ denote the shared components, and $\mathbf{p}_f,\mathbf{p}_n \in \mathbb{R}^{D_p}$ denote the private components. In this work, we set $D_s = D_p$ so that all components can be represented as tokens with the same dimensionality during cross-attention fusion. For a vector $\mathbf{v}$, we write $\bar{\mathbf{v}} = \mathbf{v}/(\|\mathbf{v}\|_2+\epsilon)$ for its $\ell_2$-normalized form.

To align the shared components of the two modalities, we define the similarity loss as the mean squared difference between the normalized shared components:
\begin{equation}
    \mathcal{L}_{\mathrm{sim}}
    =
    \frac{1}{B D_s}
    \sum_{i=1}^{B}
    \sum_{q=1}^{D_s}
    \left(
    \bar{s}_{f,q}^{(i)}
    -
    \bar{s}_{n,q}^{(i)}
    \right)^{2},
    \label{eq:sim_loss}
\end{equation}
where $B$ is the batch size and $q$ indexes the dimensions of the shared space. Each shared component is $\ell_2$-normalized before the comparison, so that the loss measures the difference in direction rather than magnitude, and the average is taken over both the batch and the feature dimensions.

To reduce redundancy between the shared and private components within each modality, we define the orthogonality loss as
\begin{equation}
    \mathcal{L}_{\mathrm{orth}}
    =
    \frac{1}{B}
    \sum_{i=1}^{B}
    \left(
    \left|
    \mathbf{s}_f^{(i)\top}
    \mathbf{p}_f^{(i)}
    \right|
    +
    \left|
    \mathbf{s}_n^{(i)\top}
    \mathbf{p}_n^{(i)}
    \right|
    \right),
    \label{eq:orth_loss}
\end{equation}
where the inner products are computed on the unnormalized components. The absolute inner products encourage the shared and private components of the same modality to be close to orthogonal.

To encourage the private components of the two modalities to remain distinct, we define the difference loss using a hinge on the mean squared distance between the normalized private components:
\begin{equation}
    \mathcal{L}_{\mathrm{diff}}
    =
    \max
    \!\left(
    0,\;
    m
    -
    \frac{1}{B}
    \sum_{i=1}^{B}
    \left\|
    \bar{\mathbf{p}}_f^{(i)}
    -
    \bar{\mathbf{p}}_n^{(i)}
    \right\|_2^{2}
    \right),
    \label{eq:diff_loss}
\end{equation}
where $m$ is the margin. The loss becomes zero once the average squared distance between the two normalized private components reaches the margin, so that the term acts as a separation constraint rather than a continuous repulsion. Together, these auxiliary losses regularize the decomposition by promoting cross-modal agreement in the shared space and separation in the private space.

\subsection{Cross-attention fusion}
\label{sec:cross_attention_fusion}

The four decomposed components are integrated through a bidirectional cross-attention mechanism. The shared and private components of each modality are first stacked into two-token sequences:
\begin{equation}
    \mathbf{Z}_f = [\mathbf{s}_f,\mathbf{p}_f], \;\;
    \mathbf{Z}_n = [\mathbf{s}_n,\mathbf{p}_n]
    \in \mathbb{R}^{2 \times D_s},
    \label{eq:token_stack}
\end{equation}
where each modality is represented by two tokens, corresponding to the shared token and the private token. At each of the $L_a$ cross-attention layers, the fMRI tokens first attend to the non-imaging tokens using multi-head attention (MHA), where the fMRI stream provides the queries and the non-imaging stream provides the keys and values. The non-imaging tokens then attend to the updated fMRI tokens in the reverse direction. Each cross-attention operation is followed by a residual connection, layer normalization (LN), and a position-wise feedforward network (FFN). The cross-attention layer is written as
\begin{equation}
\begin{aligned}
    \mathbf{Z}_f'
    &=
    \mathrm{LN}
    \left(
    \mathbf{Z}_f
    +
    \mathrm{MHA}
    (
    \mathbf{Z}_f,
    \mathbf{Z}_n,
    \mathbf{Z}_n
    )
    \right),
    \\[2pt]
    \hat{\mathbf{Z}}_f
    &=
    \mathrm{LN}
    \left(
    \mathbf{Z}_f'
    +
    \mathrm{FFN}
    (
    \mathbf{Z}_f'
    )
    \right),
    \\[2pt]
    \mathbf{Z}_n'
    &=
    \mathrm{LN}
    \left(
    \mathbf{Z}_n
    +
    \mathrm{MHA}
    (
    \mathbf{Z}_n,
    \hat{\mathbf{Z}}_f,
    \hat{\mathbf{Z}}_f
    )
    \right),
    \\[2pt]
    \hat{\mathbf{Z}}_n
    &=
    \mathrm{LN}
    \left(
    \mathbf{Z}_n'
    +
    \mathrm{FFN}
    (
    \mathbf{Z}_n'
    )
    \right),
\end{aligned}
    \label{eq:cross_attention}
\end{equation}
where the three arguments of MHA correspond to the query, key, and value, respectively.

After $L_a$ layers of cross-attention, the tokens from both modality streams are flattened and passed through a modality gate. The gate is implemented as a two-layer MLP followed by a softmax over the two modality scores. The gate-weighted outputs are concatenated and projected to produce the final fused representation:
\begin{equation}
\begin{aligned}
    \boldsymbol{\alpha}
    &=
    \mathrm{softmax}
    \left(
    \mathrm{MLP}_{g}
    \left(
    [
    \hat{\mathbf{Z}}_f^{\mathrm{flat}}
    \,\|\,
    \hat{\mathbf{Z}}_n^{\mathrm{flat}}
    ]
    \right)
    \right),
    \\[2pt]
    \mathbf{h}_{\mathrm{fused}}
    &=
    \mathrm{MLP}_{o}
    \left(
    \alpha_1
    \hat{\mathbf{Z}}_f^{\mathrm{flat}}
    \,\|\,
    \alpha_2
    \hat{\mathbf{Z}}_n^{\mathrm{flat}}
    \right),
\end{aligned}
    \label{eq:fused}
\end{equation}
where $\hat{\mathbf{Z}}_f^{\mathrm{flat}}$ and $\hat{\mathbf{Z}}_n^{\mathrm{flat}}$ denote the flattened token sequences, $\boldsymbol{\alpha}=[\alpha_1,\alpha_2]$ denotes the modality gate weights for the imaging and non-imaging streams, and $\mathrm{MLP}_{o}$ is the output projection. The modality gate assigns sample-specific weights to the two modalities before final prediction.

\subsection{Classification and cross-site supervised contrastive learning}
\label{sec:classification_contrastive}

The fused representation $\mathbf{h}_{\mathrm{fused}}$ is passed through a three-layer classification MLP with GELU activation and dropout to produce the diagnostic logits. The classification loss is computed using cross-entropy (CE) with label smoothing:
\begin{equation}
    \mathcal{L}_{\mathrm{CE}}
    =
    \mathrm{CE}
    (
    \hat{\mathbf{y}},
    \mathbf{y};
    \eta
    ),
    \label{eq:ce_loss}
\end{equation}
where $\hat{\mathbf{y}}$ is the predicted distribution, $\mathbf{y}$ is the ground-truth label, and $\eta$ is the label smoothing parameter. To address class imbalance across sites, inverse-frequency class weights are computed from the source-training set of each fold and applied to the CE loss.

The CE loss encourages correct classification but does not control how the fused representations are arranged relative to one another. In a multi-site setting, subjects from the same site share acquisition conditions such as scanner hardware and imaging protocol, and these shared conditions may cause representations to cluster by site rather than by diagnostic label. Standard supervised contrastive learning treats every same-class subject as a positive, including same-class subjects from the same site, which would draw together subjects that already share site-specific characteristics and reinforce site-dependent structure~\cite{khosla2020supervised}. To address this limitation, we propose a cross-site variant of supervised contrastive learning that operates on the fused representation and restricts the positive set to same-class subjects from different source sites.

\begin{figure}
    \centering
    \includegraphics[width=\columnwidth]{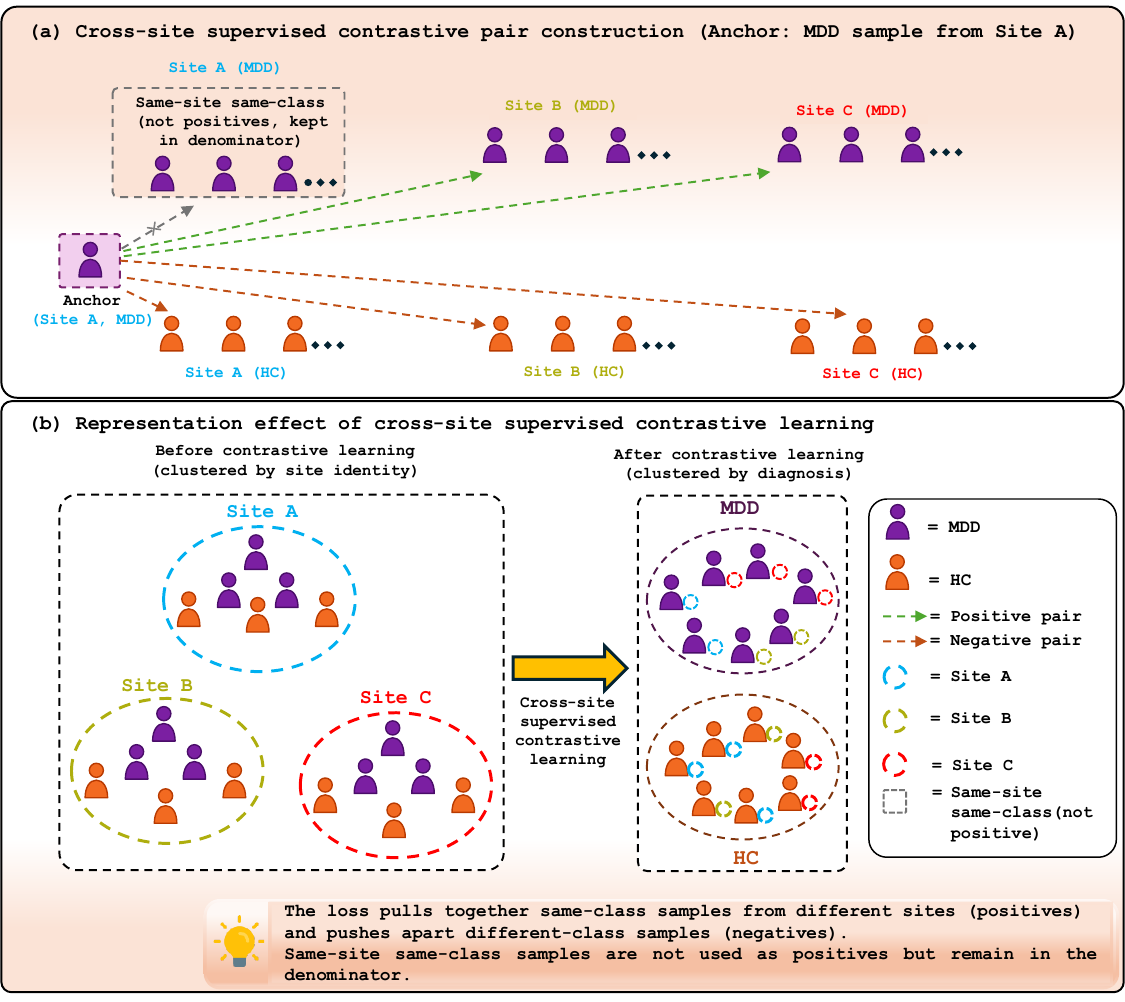}
    \caption{Illustration of cross-site supervised contrastive learning.
    (a) Pair construction for an MDD anchor from Site~A. Same-class subjects
    from different source sites form the positive set, whereas HC subjects are
    treated as negatives regardless of site. Same-site same-class subjects are
    excluded from the positive set, although the denominator of the loss
    includes all non-anchor samples in the batch.
    (b) Conceptual effect of the proposed formulation on the representation space. Before contrastive learning, subjects may form site-dependent
    clusters. After contrastive learning, same-class subjects from different
    sites are encouraged to form diagnosis-related groups with greater mixing
    of site identities.}
    \label{FIG:contrastive}
\end{figure}

Fig.~\ref{FIG:contrastive}(a) illustrates the pair construction for an MDD anchor from Site~A. MDD subjects from other sites form positive pairs, subjects with a different diagnostic label are treated as negatives regardless of site, and MDD subjects from Site~A are excluded from the positive set but remain in the denominator. For each anchor subject $i$ in a training batch, the positive set and the normalized representation are defined as
\begin{equation}
\begin{aligned}
    \mathcal{P}(i)
    &=
    \{\,j \neq i : y_j=y_i,\; \omega_j \neq \omega_i\,\},
    \\[2pt]
    \mathbf{z}_i
    &=
    \frac{\mathbf{h}_{\mathrm{fused}}^{(i)}}
    {\|\mathbf{h}_{\mathrm{fused}}^{(i)}\|_2+\epsilon},
\end{aligned}
    \label{eq:positive_set}
\end{equation}
where $\mathcal{P}(i)$ is the positive set of anchor $i$, $j$ indexes the other subjects in the batch, and $\mathbf{z}_i$ is the $\ell_2$-normalized fused representation. The cross-site supervised contrastive loss is then defined as
\begin{equation}
    \mathcal{L}_{\mathrm{con}}
    =
    -
    \frac{1}{|\mathcal{V}|}
    \sum_{i \in \mathcal{V}}
    \frac{1}{|\mathcal{P}(i)|}
    \sum_{j \in \mathcal{P}(i)}
    \log
    \frac{
    \exp
    (
    \mathbf{z}_i^\top
    \mathbf{z}_j
    /
    \tau
    )
    }{
    \displaystyle
    \sum_{k \neq i}
    \exp
    (
    \mathbf{z}_i^\top
    \mathbf{z}_k
    /
    \tau
    )
    },
    \label{eq:contrastive}
\end{equation}
where $\tau$ is the temperature parameter, and $\mathcal{V}=\{\,i:|\mathcal{P}(i)|>0\,\}$ is the set of anchors that have at least one valid cross-site same-class positive. The denominator sums over all non-self samples in the batch, so same-site same-class samples remain in the denominator but are not treated as positives. As a result, same-class different-site samples receive an attractive effect through the positive term, while same-class same-site samples receive only an indirect repulsive effect through the denominator. This site-aware treatment reduces the influence of similarities that may arise from shared acquisition conditions within the same site and places greater emphasis on diagnostic consistency across sites, as illustrated conceptually in Fig.~\ref{FIG:contrastive}(b). Anchors without a valid cross-site same-class positive are excluded from $\mathcal{L}_{\mathrm{con}}$ but still contribute to the classification loss. During training, mini-batches are randomly sampled from the complete source-training set, which contains subjects from multiple source sites and both diagnostic classes whenever available.

The fused representation is simultaneously optimized for classification and for cross-site consistency. The classification loss $\mathcal{L}_{\mathrm{CE}}$ is applied to the logits produced by the classification head, while the cross-site supervised contrastive loss $\mathcal{L}_{\mathrm{con}}$ is applied directly to the $\ell_2$-normalized fused representation. Both terms are combined in the total objective in Eq.~\eqref{eq:total_loss} and optimized jointly.

\subsection{Optimization}
\label{sec:optimization}

The total training objective combines the classification loss, the three
decomposition losses, and the cross-site supervised contrastive loss:
\begin{equation}
\resizebox{\columnwidth}{!}{$
\displaystyle
\mathcal{L}_{\mathrm{total}}
=
\mathcal{L}_{\mathrm{CE}}
+\lambda_{\mathrm{sim}}\mathcal{L}_{\mathrm{sim}}
+\lambda_{\mathrm{orth}}\mathcal{L}_{\mathrm{orth}}
+\lambda_{\mathrm{diff}}\mathcal{L}_{\mathrm{diff}}
+\lambda_{\mathrm{con}}\mathcal{L}_{\mathrm{con}}
$}
\label{eq:total_loss}
\end{equation}
where $\lambda_{\mathrm{sim}}$, $\lambda_{\mathrm{orth}}$,
$\lambda_{\mathrm{diff}}$, and $\lambda_{\mathrm{con}}$ control the relative
contributions of the corresponding auxiliary losses. To reduce the number of
hyperparameters, the three decomposition losses share a single weight
$\lambda_{\mathrm{decomp}}$:
\begin{equation}
\lambda_{\mathrm{sim}}
=
\lambda_{\mathrm{orth}}
=
\lambda_{\mathrm{diff}}
=
\lambda_{\mathrm{decomp}}.
\label{eq:decomp_weight}
\end{equation}

\section{Experiments and results}
\label{sec:exp_result}

\subsection{Experimental setup}
\subsubsection{Datasets}
\label{sec:Data_MDD}

The REST-meta-MDD Project, established under the Depression Imaging REsearch ConsorTium (DIRECT), is a large-scale publicly available dataset for MDD research. DIRECT was initiated in 2017, and in 2020, twenty-five research groups from seventeen hospitals across China contributed rs-fMRI data from patients with MDD and matched HCs as the consortium's first collaborative project~\cite{yan2019reduced,chen2022direct}. The full dataset contains rs-fMRI recordings from 1,300 patients with MDD and 1,128 HCs, and it has been widely adopted in recent MDD studies, providing a common benchmark for fair comparison and cross-validation of findings~\cite{qiu2025metaexplainer,su2024m2dc}. A standardized preprocessing pipeline adapted from the Data Processing Assistant for rs-fMRI (DPARSF) was applied at each participating site~\cite{yan2016dpabi,yan2010dparsf}. The first ten volumes were discarded to allow for signal stabilization, followed by slice-timing correction and realignment using a six-parameter rigid-body linear transformation. Individual T1-weighted images were coregistered to the mean functional image without resampling and segmented into gray matter, white matter, and cerebrospinal fluid~\cite{ashburner2005unified}. Transformations from native space to Montreal Neurological Institute (MNI) space were then computed using DARTEL~\cite{ashburner2007fast}. Head motion effects were regressed out using the Friston 24-parameter model~\cite{friston1996movement}, while global signal regression was not performed. White matter signals, cerebrospinal fluid signals, and linear trends were additionally removed through linear regression, and temporal band-pass filtering at 0.01--0.1~Hz was applied to all time series. Following prior work~\cite{su2024m2dc}, Sites~5 and~19 were excluded from the dataset because they did not provide the time courses required for connectome construction. Site~4 was further excluded because its data duplicated those from Site~14, thereby preventing data leakage during performance evaluation~\cite{qiu2025metaexplainer}. After these exclusions, the final dataset used in this study comprised 1,212 patients with MDD and 1,057 HCs distributed across 22 sites. Whole-brain parcellation was performed using the Automated Anatomical Labeling (AAL) atlas, yielding $N = 116$ ROIs~\cite{tzourio2002automated}. 

\begin{table}[width=.9\linewidth,cols=2,pos=h]
\caption{Hyperparameters for the proposed M$^2$LG-DG framework.}\label{tbl:hyperparams}
\begin{tabular*}{\tblwidth}{@{} LL@{} }
\toprule
\multicolumn{2}{@{}l}{\textbf{Dual-Stream fMRI Encoder}} \\
\midrule
Global encoder Conv1D channels & [32, 64, 128] \\
Transformer hidden dimension & 128 \\
Transformer attention heads & 8 \\
Transformer layers & 2 \\
Global encoder output dimension & 256 \\
Local encoder ST-GCN channels & [64, 128, 256] \\
ST-GCN temporal kernel size & 9 \\
Chebyshev order $K$ & 3 \\
Local encoder output dimension & 256 \\
\midrule
\multicolumn{2}{@{}l}{\textbf{Non-imaging Encoder}} \\
\midrule
MLP hidden dimensions & [32, 64] \\
Non-imaging output dimension & 64 \\
\midrule
\multicolumn{2}{@{}l}{\textbf{Shared-Private Decomposition \& Fusion}} \\
\midrule
Shared and private dimensions & 128 \\
Cross-attention heads & 4 \\
Cross-attention layers & 2 \\
Fusion output dimension & 128 \\
\midrule
\multicolumn{2}{@{}l}{\textbf{Training Configuration}} \\
\midrule
Optimizer & AdamW \\
Learning rate & $1 \times 10^{-4}$ \\
Weight decay & $1 \times 10^{-4}$ \\
Learning rate scheduler & Cosine annealing \\
Warmup epochs & 10 \\
Batch size & 32 \\
Maximum epochs & 100 \\
Early stopping patience & 20 \\
Gradient clipping & 1.0 \\
Dropout & 0.1 \\
\midrule
\multicolumn{2}{@{}l}{\textbf{Loss Configuration}} \\
\midrule
Label smoothing $\eta$ & 0.1 \\
Contrastive temperature $\tau$ & 0.07 \\
Decomposition loss weight $\lambda_{\text{decomp}}$ & 0.1 \\
Contrastive loss weight $\lambda_{\text{con}}$ & 0.1 \\
\bottomrule
\end{tabular*}
\end{table}

\subsubsection{Implementation details}
\label{sec:implementation}
The evaluation was conducted under a strict DG setting, in which each target domain remained entirely unseen during model development. For the primary evaluation, four sites were considered as unseen target domains: Site~9 (100 subjects), Site~15 (100 subjects), Site~20 (533 subjects), and Site~21 (156 subjects). These sites were selected to represent target domains with cohort sizes ranging from 100 to 533 subjects, allowing performance to be assessed under different target-sample conditions. The target sites were selected based only on participant counts, before model training and before any target-site performance was examined. For each evaluation, one site was designated as the target domain and used exclusively for testing, while samples from all remaining sites in the REST-meta-MDD project were used for model development. This procedure was repeated using five random seeds, and the performance for each held-out target site was reported as the mean and standard deviation across the five runs. Overall performance was calculated by averaging the per-site mean values across the four target sites. Model performance was evaluated using accuracy (ACC), sensitivity (SEN), specificity (SPE), F1-score (F1), and the area under the receiver operating characteristic curve (AUC).

All experiments were conducted on an NVIDIA RTX 4090 GPU using PyTorch~2.1. The rs-fMRI time series were interpolated to a fixed length of $T_{\mathrm{fixed}} = 200$ time points and z-score normalized for each ROI within each subject. The group-level FC matrix was computed from training subjects only and thresholded at the 80th percentile to construct the adjacency matrix, with self-loops added. The architectural and training hyperparameters are summarized in Table~\ref{tbl:hyperparams}. Sensitivity analyses for $\lambda_{\text{decomp}}$, $\lambda_{\text{con}}$, $\tau$, the FC threshold percentile, and $T_{\mathrm{fixed}}$ are presented in Section~\ref{sec:sensitivity_study}.

\begin{table*}[t]
\centering
\caption{Classification performance (\%) on the REST-meta-MDD dataset, reported as the mean $\pm$ standard deviation over five random seeds. Site~9, Site~15, Site~20, and Site~21 were each treated as a held-out target domain in a separate experiment, while the
remaining 21 of the 22 included sites were used as source domains for model
development. The best result in each column is shown in bold and the second-best result is
underlined.}
\label{tab:internal_validation_rest_meta_mdd}
\small
\setlength{\tabcolsep}{3.5pt}
\renewcommand{\arraystretch}{1.10}

\resizebox{\textwidth}{!}{%
\begin{tabular}{l *{10}{c}}
\toprule
\multicolumn{11}{c}{\textbf{REST-meta-MDD dataset}} \\
\midrule
& \multicolumn{5}{c}{Site 9}
& \multicolumn{5}{c}{Site 15} \\
\cmidrule(lr){2-6} \cmidrule(lr){7-11}
Method
& {ACC} & {SPE} & {SEN} & {AUC} & {F1}
& {ACC} & {SPE} & {SEN} & {AUC} & {F1} \\
\midrule
BrainNetCNN          & 53.40 $\pm$ 3.05 & 62.90 $\pm$ 10.84 & 44.60 $\pm$ 9.12 & 60.30 $\pm$ 6.41 & 41.78 $\pm$ 7.20 & 54.90 $\pm$ 3.48 & 58.10 $\pm$ 8.22 & 51.80 $\pm$ 10.10 & 58.70 $\pm$ 5.55 & 53.48 $\pm$ 6.90 \\
CDANN                & 54.60 $\pm$ 2.74 & 64.80 $\pm$ 13.10 & \underline{44.90 $\pm$ 12.90} & 63.20 $\pm$ 9.55 & 42.71 $\pm$ 5.12 & 56.40 $\pm$ 2.58 & 59.70 $\pm$ 6.02 & 53.20 $\pm$ 7.05 & 60.80 $\pm$ 8.44 & 54.99 $\pm$ 4.60 \\
Mixup                & 55.10 $\pm$ 4.02 & 66.40 $\pm$ 6.88 & 43.70 $\pm$ 7.95 & 62.10 $\pm$ 6.02 & 42.41 $\pm$ 11.40 & 53.20 $\pm$ 4.51 & 58.90 $\pm$ 4.05 & 47.80 $\pm$ 7.70 & 58.20 $\pm$ 3.60 & 50.61 $\pm$ 11.80 \\
MLDG                 & 54.90 $\pm$ 5.55 & 65.10 $\pm$ 15.10 & 44.20 $\pm$ 15.40 & 60.70 $\pm$ 8.62 & 42.29 $\pm$ 12.80 & 55.40 $\pm$ 5.44 & 58.40 $\pm$ 7.44 & 52.30 $\pm$ 15.80 & 59.50 $\pm$ 6.44 & 53.95 $\pm$ 10.85 \\
BrainGNN             & 56.70 $\pm$ 4.28 & 70.80 $\pm$ 9.85 & 42.90 $\pm$ 8.95 & 64.10 $\pm$ 5.98 & 43.52 $\pm$ 6.20 & 53.90 $\pm$ 2.02 & 60.30 $\pm$ 15.30 & 47.40 $\pm$ 15.02 & 57.60 $\pm$ 10.70 & 50.67 $\pm$ 10.12 \\
IIB                  & 53.60 $\pm$ 4.24 & 62.40 $\pm$ 14.30 & 44.10 $\pm$ 5.75 & 63.40 $\pm$ 8.08 & 41.23 $\pm$ 7.52 & 56.70 $\pm$ 2.30 & 60.50 $\pm$ 6.70 & 52.90 $\pm$ 11.15 & 60.90 $\pm$ 4.50 & 54.99 $\pm$ 7.72 \\
IBGNN+               & 56.90 $\pm$ 4.50 & 69.00 $\pm$ 9.40 & 44.60 $\pm$ 5.70 & 67.40 $\pm$ 6.51 & 44.12 $\pm$ 5.36 & 55.80 $\pm$ 2.61 & 61.40 $\pm$ 6.58 & 50.10 $\pm$ 8.96 & 66.10 $\pm$ 2.98 & 53.10 $\pm$ 9.74 \\
UFA-Net              & 55.40 $\pm$ 6.28 & 67.10 $\pm$ 9.34 & 43.70 $\pm$ 13.90 & 65.20 $\pm$ 10.24 & 42.68 $\pm$ 3.38 & 54.30 $\pm$ 2.12 & 60.70 $\pm$ 15.10 & 48.00 $\pm$ 15.55 & 58.60 $\pm$ 7.77 & 51.25 $\pm$ 4.55 \\
GenM                 & 56.10 $\pm$ 3.52 & 69.60 $\pm$ 15.98 & 42.40 $\pm$ 16.22 & 62.40 $\pm$ 9.14 & 42.64 $\pm$ 5.02 & 57.40 $\pm$ 5.78 & 61.70 $\pm$ 9.08 & 53.10 $\pm$ 13.85 & 61.50 $\pm$ 3.50 & 55.49 $\pm$ 13.06 \\
MSSTAN               & 61.40 $\pm$ 3.12 & 77.90 $\pm$ 13.66 & 44.70 $\pm$ 11.52 & 69.50 $\pm$ 5.60 & 48.13 $\pm$ 7.76 & 63.90 $\pm$ 3.62 & 61.20 $\pm$ 7.64 & 66.60 $\pm$ 10.84 & 65.60 $\pm$ 5.30 & 64.85 $\pm$ 8.50 \\
M$^2$DC              & 63.20 $\pm$ 3.24 & 81.40 $\pm$ 14.72 & 44.10 $\pm$ 7.40 & 72.00 $\pm$ 6.70 & 49.37 $\pm$ 6.26 & 64.80 $\pm$ 3.12 & 61.30 $\pm$ 8.02 & \underline{68.30 $\pm$ 10.28} & 66.40 $\pm$ 4.60 & 65.99 $\pm$ 7.00 \\
MM-GTUNets           & \underline{63.70 $\pm$ 3.48} & \underline{82.70 $\pm$ 12.90} & \underline{44.90 $\pm$ 8.70} & \underline{73.60 $\pm$ 4.98} & \underline{50.73 $\pm$ 7.34} & \underline{65.20 $\pm$ 3.00} & \textbf{62.50 $\pm$ 7.34} & 67.90 $\pm$ 10.58 & 66.30 $\pm$ 4.72 & \underline{66.11 $\pm$ 7.44} \\
MetaExplainer        & 63.10 $\pm$ 2.86 & 81.90 $\pm$ 14.22 & 44.40 $\pm$ 7.02 & 72.60 $\pm$ 5.33 & 49.88 $\pm$ 6.06 & 64.40 $\pm$ 3.25 & 61.70 $\pm$ 8.40 & 68.10 $\pm$ 9.78 & \underline{66.50 $\pm$ 4.42} & 65.99 $\pm$ 6.44 \\
\midrule
\textbf{M$^2$LG-DG}  & \textbf{65.80 $\pm$ 2.93} & \textbf{85.20 $\pm$ 6.88} & \textbf{46.40 $\pm$ 11.83} & \textbf{75.81 $\pm$ 2.18} & \textbf{56.60 $\pm$ 8.82} & \textbf{66.20 $\pm$ 2.93} & \underline{62.00 $\pm$ 6.32} & \textbf{70.40 $\pm$ 5.85} & \textbf{68.40 $\pm$ 3.90} & \textbf{67.49 $\pm$ 3.11} \\
\bottomrule
\end{tabular}%
}

\vspace{0.75em}

\resizebox{\textwidth}{!}{%
\begin{tabular}{l *{10}{c}}
\toprule
& \multicolumn{5}{c}{Site 20}
& \multicolumn{5}{c}{Site 21} \\
\cmidrule(lr){2-6} \cmidrule(lr){7-11}
Method
& {ACC} & {SPE} & {SEN} & {AUC} & {F1}
& {ACC} & {SPE} & {SEN} & {AUC} & {F1} \\
\midrule
BrainNetCNN          & 56.80 $\pm$ 3.20 & 61.40 $\pm$ 10.55 & 52.90 $\pm$ 9.44 & 60.10 $\pm$ 6.28 & 56.12 $\pm$ 5.90 & 43.20 $\pm$ 3.55 & 33.80 $\pm$ 8.44 & 50.90 $\pm$ 10.30 & 58.40 $\pm$ 5.62 & 52.20 $\pm$ 6.55 \\
CDANN                & 57.50 $\pm$ 2.90 & 62.10 $\pm$ 13.44 & 53.60 $\pm$ 13.60 & 63.90 $\pm$ 9.60 & 56.84 $\pm$ 4.84 & 44.80 $\pm$ 2.72 & 34.90 $\pm$ 6.30 & 52.70 $\pm$ 7.44 & 59.60 $\pm$ 8.60 & 53.75 $\pm$ 4.28 \\
Mixup                & 55.70 $\pm$ 4.18 & 60.80 $\pm$ 6.95 & 51.10 $\pm$ 8.06 & 61.40 $\pm$ 6.14 & 54.57 $\pm$ 11.60 & 42.60 $\pm$ 4.55 & 35.20 $\pm$ 4.10 & 47.90 $\pm$ 7.85 & 58.10 $\pm$ 3.66 & 50.13 $\pm$ 12.00 \\
MLDG                 & 57.20 $\pm$ 5.60 & 61.60 $\pm$ 15.20 & 53.10 $\pm$ 15.55 & 60.60 $\pm$ 8.74 & 56.33 $\pm$ 12.95 & 44.10 $\pm$ 5.50 & 34.40 $\pm$ 7.52 & 52.10 $\pm$ 15.90 & 59.20 $\pm$ 6.52 & 53.21 $\pm$ 10.95 \\
BrainGNN             & 54.60 $\pm$ 4.32 & 61.90 $\pm$ 9.98 & 48.70 $\pm$ 9.05 & 63.80 $\pm$ 6.02 & 52.97 $\pm$ 6.28 & 41.50 $\pm$ 2.02 & 35.60 $\pm$ 15.44 & 46.20 $\pm$ 15.14 & 57.30 $\pm$ 10.85 & 48.85 $\pm$ 10.22 \\
IIB                  & 58.20 $\pm$ 4.28 & 62.00 $\pm$ 14.42 & 55.30 $\pm$ 5.82 & 63.10 $\pm$ 8.14 & 58.10 $\pm$ 7.60 & 44.90 $\pm$ 2.32 & 36.50 $\pm$ 6.75 & 51.40 $\pm$ 11.25 & 60.20 $\pm$ 4.55 & 53.06 $\pm$ 7.80 \\
IBGNN+               & 59.70 $\pm$ 4.55 & 62.30 $\pm$ 9.50 & 57.40 $\pm$ 5.75 & 66.50 $\pm$ 6.58 & 59.73 $\pm$ 5.42 & 44.60 $\pm$ 2.64 & \textbf{38.70 $\pm$ 6.62} & 49.30 $\pm$ 9.04 & 60.30 $\pm$ 3.00 & 51.85 $\pm$ 9.86 \\
UFA-Net              & 58.60 $\pm$ 6.35 & 62.40 $\pm$ 9.44 & 55.10 $\pm$ 14.05 & 65.30 $\pm$ 10.32 & 58.06 $\pm$ 3.42 & 42.90 $\pm$ 2.14 & 36.10 $\pm$ 15.24 & 47.80 $\pm$ 15.70 & 58.40 $\pm$ 7.84 & 50.21 $\pm$ 4.60 \\
GenM                 & 55.80 $\pm$ 3.55 & 62.70 $\pm$ 16.14 & 49.60 $\pm$ 16.40 & 62.40 $\pm$ 9.22 & 53.90 $\pm$ 5.08 & 45.70 $\pm$ 5.82 & 37.00 $\pm$ 9.16 & 52.90 $\pm$ 14.02 & 61.50 $\pm$ 3.52 & 54.28 $\pm$ 13.20 \\
MSSTAN               & 62.10 $\pm$ 3.34 & 62.30 $\pm$ 12.40 & 61.90 $\pm$ 13.42 & 66.70 $\pm$ 5.20 & 62.94 $\pm$ 7.44 & 55.30 $\pm$ 2.94 & 36.40 $\pm$ 11.66 & 71.10 $\pm$ 13.05 & 61.70 $\pm$ 5.80 & 66.60 $\pm$ 7.68 \\
M$^2$DC              & 61.80 $\pm$ 3.66 & \underline{63.10 $\pm$ 9.05} & 60.70 $\pm$ 11.78 & 67.20 $\pm$ 4.85 & 62.33 $\pm$ 6.70 & 56.60 $\pm$ 2.82 & 37.10 $\pm$ 8.30 & 72.40 $\pm$ 11.25 & \underline{62.00 $\pm$ 4.58} & 67.56 $\pm$ 7.22 \\
MM-GTUNets           & 62.40 $\pm$ 3.38 & 62.80 $\pm$ 10.88 & 61.90 $\pm$ 12.35 & 67.60 $\pm$ 4.55 & 63.09 $\pm$ 6.90 & \underline{56.80 $\pm$ 2.90} & 37.40 $\pm$ 8.86 & 72.60 $\pm$ 11.58 & 61.70 $\pm$ 4.90 & \underline{67.74 $\pm$ 7.00} \\
MetaExplainer        & \underline{62.50 $\pm$ 3.22} & 62.60 $\pm$ 10.18 & \underline{62.10 $\pm$ 11.00} & \underline{68.00 $\pm$ 4.20} & \underline{63.17 $\pm$ 6.28} & 56.40 $\pm$ 3.15 & 36.80 $\pm$ 9.62 & \underline{73.00 $\pm$ 10.52} & 61.90 $\pm$ 4.70 & \textbf{67.86 $\pm$ 6.66} \\
\midrule
\textbf{M$^2$LG-DG}  & \textbf{64.24 $\pm$ 2.56} & \textbf{64.78 $\pm$ 12.84} & \textbf{63.76 $\pm$ 16.05} & \textbf{70.16 $\pm$ 0.91} & \textbf{64.27 $\pm$ 7.83} & \textbf{58.33 $\pm$ 1.72} & \underline{37.71 $\pm$ 17.06} & \textbf{75.12 $\pm$ 12.60} & \textbf{63.53 $\pm$ 1.90} & 66.12 $\pm$ 3.73 \\
\bottomrule
\end{tabular}%
}

\end{table*}

\begin{table*}[t]
\centering
\caption{Classification performance (\%) on the REST-meta-MDD dataset across
the four held-out target sites (Site~9, Site~15, Site~20, and Site~21).
Panel~A reports the unweighted macro-average performance across the four
target sites, while Panel~B reports paired statistical comparisons between
M$^{2}$LG-DG and four competitive baseline methods.}
\label{tab:overall_rest_meta_mdd}
\footnotesize
\renewcommand{\arraystretch}{1.0}
\setlength{\tabcolsep}{6pt}

\begin{tabular}{@{}lccccc@{}}
\toprule
\multicolumn{6}{c}{\underline{\textit{Panel A: Overall performance across the four held-out target sites}}}\\[1pt]
\textbf{Method} & \textbf{ACC} & \textbf{SPE} & \textbf{SEN} & \textbf{AUC} & \textbf{F1} \\
\midrule
BrainNetCNN
& $52.08\pm6.08$
& $54.05\pm13.65$
& $50.05\pm3.72$
& $59.38\pm0.96$
& $50.89\pm6.29$ \\

CDANN
& $53.33\pm5.81$
& $55.38\pm13.81$
& $51.10\pm4.15$
& $61.88\pm2.02$
& $52.07\pm6.37$ \\

Mixup
& $51.65\pm6.13$
& $55.33\pm13.79$
& $47.62\pm3.03$
& $59.95\pm2.10$
& $49.43\pm5.09$ \\

MLDG
& $52.90\pm5.95$
& $54.88\pm13.92$
& $50.42\pm4.17$
& $60.00\pm0.76$
& $51.45\pm6.25$ \\

BrainGNN
& $51.67\pm6.89$
& $57.15\pm15.09$
& $46.30\pm2.49$
& $60.70\pm3.76$
& $49.00\pm4.03$ \\

IIB
& $53.35\pm5.95$
& $55.35\pm12.59$
& $50.93\pm4.83$
& $61.90\pm1.59$
& $51.84\pm7.37$ \\

IBGNN+
& $54.25\pm6.64$
& $57.85\pm13.21$
& $50.35\pm5.29$
& $65.08\pm3.23$
& $52.20\pm6.40$ \\

UFA-Net
& $52.80\pm6.85$
& $56.57\pm13.92$
& $48.65\pm4.73$
& $61.88\pm3.90$
& $50.55\pm6.30$ \\

GenM
& $53.75\pm5.41$
& $57.75\pm14.27$
& $49.50\pm5.00$
& $61.95\pm0.52$
& $51.58\pm6.00$ \\

MSSTAN
& $60.67\pm3.73$
& $59.45\pm17.16$
& $61.07\pm11.54$
& $65.88\pm3.23$
& $60.63\pm8.47$ \\

M$^{2}$DC
& $61.60\pm3.55$
& $60.72\pm18.18$
& $61.38\pm12.50$
& $66.90\pm4.10$
& $61.31\pm8.26$ \\

MM-GTUNets
& $\underline{62.03\pm3.67}$
& $\underline{61.35\pm18.55}$
& $61.83\pm12.10$
& $\underline{67.30\pm4.90}$
& $\underline{61.92\pm7.70}$ \\

MetaExplainer
& $61.60\pm3.56$
& $60.75\pm18.49$
& $\underline{61.90\pm12.49}$
& $67.25\pm4.41$
& $61.73\pm8.13$ \\

\midrule
\textbf{M$^{2}$LG-DG}
& $\mathbf{63.64\pm3.64}$
& $\mathbf{62.42\pm19.45}$
& $\mathbf{63.92\pm12.58}$
& $\mathbf{69.48\pm5.07}$
& $\mathbf{63.62\pm4.86}$ \\
\bottomrule
\end{tabular}

\vspace{1.5mm}

\setlength{\tabcolsep}{2.5pt}
\begin{tabular*}{\textwidth}{@{\extracolsep{\fill}}l ccc ccc ccc ccc@{}}
\toprule
\multicolumn{13}{c}{\underline{\textit{Panel B: Paired statistical comparisons relative to M$^{2}$LG-DG}}}\\[1pt]

& \multicolumn{3}{c}{\textbf{ACC}}
& \multicolumn{3}{c}{\textbf{SEN}}
& \multicolumn{3}{c}{\textbf{AUC}}
& \multicolumn{3}{c}{\textbf{F1}} \\

\cmidrule(lr){2-4}
\cmidrule(lr){5-7}
\cmidrule(lr){8-10}
\cmidrule(lr){11-13}

\textbf{Method}
& $\Delta$ & $t$ & $p_{\mathrm{Holm}}$
& $\Delta$ & $t$ & $p_{\mathrm{Holm}}$
& $\Delta$ & $t$ & $p_{\mathrm{Holm}}$
& $\Delta$ & $t$ & $p_{\mathrm{Holm}}$ \\
\midrule

\textbf{M$^{2}$LG-DG}
& $0.00$ & -- & --
& $0.00$ & -- & --
& $0.00$ & -- & --
& $0.00$ & -- & -- \\

\midrule

MSSTAN
& $+2.97$ & $5.76$ & $0.017^{*}$
& $+2.85$ & $4.61$ & $0.019^{*}$
& $+3.60$ & $3.74$ & $0.033^{*}$
& $+2.99$ & $1.54$ & $0.880$ \\

M$^{2}$DC
& $+2.04$ & $7.15$ & $0.017^{*}$
& $+2.54$ & $11.85$ & $0.004^{**}$
& $+2.57$ & $5.07$ & $0.030^{*}$
& $+2.31$ & $1.28$ & $0.880$ \\

MM-GTUNets
& $+1.62$ & $6.84$ & $0.017^{*}$
& $+2.09$ & $8.36$ & $0.007^{**}$
& $+2.17$ & $14.39$ & $0.003^{**}$
& $+1.70$ & $1.10$ & $0.880$ \\

MetaExplainer
& $+2.04$ & $9.17$ & $0.011^{*}$
& $+2.02$ & $14.97$ & $0.003^{**}$
& $+2.22$ & $6.44$ & $0.023^{*}$
& $+1.89$ & $1.08$ & $0.880$ \\

\bottomrule
\end{tabular*}

\vspace{1mm}

\begin{minipage}{0.97\textwidth}
\scriptsize
Per-site performance was first averaged across five random seeds. Panel~A
presents the mean $\pm$ standard deviation of the resulting site-level scores,
so the reported dispersion reflects variation across target sites rather than
across random seeds. The mean represents an unweighted macro-average in which
each site contributes equally regardless of cohort size. Bold and underlined
values denote the best and second-best results, respectively. Panel~B presents
two-sided paired $t$-tests comparing M$^2$LG-DG with MSSTAN, M$^2$DC,
MM-GTUNets, and MetaExplainer, using the four target sites as paired observations
($n=4$, $\mathrm{df}=3$). Here, $\Delta$ denotes the mean difference in
percentage points in favor of M$^2$LG-DG, and Holm correction was applied within
each metric across the four comparisons. The test therefore reflects the
consistency of the site-level differences rather than sampling variation across
individual participants. Specificity was evaluated using the same procedure and
is not shown in Panel~B because none of the comparisons reached statistical
significance ($\Delta$ ranging from 1.07 to 2.97,
$p_{\mathrm{Holm}}=0.351$), consistent with its between-site standard deviation
of 13 to 19 percentage points. $^{*}p<0.05$, $^{**}p<0.01$,
$^{***}p<0.001$.
\end{minipage}

\end{table*}

\subsection{Diagnosis performance of M$^2$LG-DG}

We compare M$^2$LG-DG with thirteen baseline methods on four held-out target
sites of the REST-meta-MDD dataset following the site-wise evaluation protocol
described in Section~\ref{sec:implementation}. In this setting, one site is
treated as the target domain, while the remaining sites are used as source
domains. The comparison methods include BrainNetCNN~\cite{kawahara2017brainnetcnn},
CDANN~\cite{li2018deep}, Interdomain Mixup (Mixup)~\cite{zhang2017mixup},
MLDG~\cite{li2018learning}, BrainGNN~\cite{li2021braingnn},
IIB~\cite{li2022invariant}, IBGNN+~\cite{cui2022interpretable},
UFA-Net~\cite{fang2023unsupervised}, GenM~\cite{lee2023site},
MSSTAN~\cite{kong2024multi}, M$^2$DC~\cite{su2024m2dc},
MM-GTUNets~\cite{cai2025mm}, and MetaExplainer~\cite{qiu2025metaexplainer}.
Among these methods, MM-GTUNets also uses non-imaging variables, whereas the
other methods rely on rs-fMRI data alone. The evaluation is performed on
Site~9, Site~15, Site~20, and Site~21 using ACC, SPE, SEN, AUC, and F1, with
each result reported as the mean and standard deviation over five random seeds.
Table~\ref{tab:internal_validation_rest_meta_mdd} reports the performance for
each held-out target site, while Table~\ref{tab:overall_rest_meta_mdd}
summarizes the macro-average results and paired statistical comparisons across
the four target sites.

As presented in Table~\ref{tab:internal_validation_rest_meta_mdd}, M$^2$LG-DG
achieves the highest values for all five evaluation metrics at Site~9 and
Site~20. At Site~9, it obtains 65.80\% ACC, 85.20\% SPE, 46.40\% SEN, 75.81\%
AUC, and 56.60\% F1, while at Site~20 the corresponding values are 64.24\%,
64.78\%, 63.76\%, 70.16\%, and 64.27\%, respectively. At Site~15,
M$^2$LG-DG achieves the highest ACC, SEN, AUC, and F1 values of 66.20\%,
70.40\%, 68.40\%, and 67.49\%, while its SPE of 62.00\% is second to the
62.50\% achieved by MM-GTUNets. At Site~21, it obtains the highest ACC, SEN,
and AUC values of 58.33\%, 75.12\%, and 63.53\%, while its SPE of 37.71\%
ranks second to IBGNN+. Its F1 value of 66.12\% at this site is lower than
those obtained by MSSTAN, M$^2$DC, MM-GTUNets, and MetaExplainer, which is
consistent with the lower specificity observed across methods at Site~21 and
the dependence of threshold-based metrics on the selected operating point.
Across the four target sites, M$^2$LG-DG ranks first in ACC, SEN, and AUC at
each site, first in F1 at Sites~9, 15, and 20, and first in SPE at Sites~9 and
20. The variation in SPE and SEN across sites is observed across the compared
methods and is consistent with differences in class composition and acquisition
conditions among the participating centers.

The macro-average results are presented in Panel~A of
Table~\ref{tab:overall_rest_meta_mdd}. M$^2$LG-DG achieves the highest values
for all five metrics, with 63.64\% ACC, 62.42\% SPE, 63.92\% SEN, 69.48\% AUC,
and 63.62\% F1. Its AUC is 2.18 percentage points higher than that of the
second-ranked MM-GTUNets. The standard deviations reported in Panel~A represent
variation across the four target domains rather than across random seeds and
are larger for SPE and SEN, which is consistent with the site-wise results
described above. Panel~B presents paired comparisons with MSSTAN, M$^2$DC,
MM-GTUNets, and MetaExplainer, with Holm correction applied separately within
each metric across the four comparisons. The differences in ACC (1.62 to 2.97
percentage points), SEN (2.02 to 2.85), and AUC (2.17 to 3.60) remain
statistically significant against all four methods after correction, whereas
the corresponding differences in F1 (1.70 to 2.99) and SPE do not reach
statistical significance. Overall, the site-wise and macro-average results show
that M$^2$LG-DG provides consistent performance in AUC, ACC, and SEN across the evaluated target domains, while SPE and F1 exhibit greater variation among the target cohorts.

\subsection{Ablation study}
\label{sec:ablation_study}

To evaluate the contribution of the input modalities and the main components
of M$^2$LG-DG, we conducted ablation experiments using the same site-wise
source-only evaluation protocol as in the main experiments. The evaluated
components include the dual-stream fMRI encoder (\textbf{Dual}), the
non-imaging encoder (\textbf{Non-img}), the shared-private decomposition
losses (\textbf{Decomp}), the bidirectional cross-attention fusion module
with the learned modality gate (\textbf{CrossAttn}), and the cross-site
supervised contrastive loss (\textbf{Contrastive}). The global and local
pathways were jointly treated as the Dual component. For each evaluation
fold, the held-out target site remained excluded from training,
preprocessing-statistic estimation, graph construction, validation, early
stopping, hyperparameter selection, and model selection.

The ablation study is organized into two groups. The first group evaluates the
contribution of each input modality independently using two configurations:
\textbf{M1}, which uses only the non-imaging variables (age, sex, and years of
education); and \textbf{M2}, which uses only the dual-stream fMRI encoder. The
second group evaluates the main components of the complete framework through a
nested sequence of four configurations: \textbf{C1}, which removes
$\mathcal{L}_{\text{sim}}$, $\mathcal{L}_{\text{orth}}$, and
$\mathcal{L}_{\text{diff}}$; \textbf{C2}, which retains the decomposition
losses but replaces the fusion module with concatenation followed by linear
projection; \textbf{C3}, which retains the complete multi-modal architecture
but removes $\mathcal{L}_{\text{con}}$; and \textbf{C4}, which represents the
complete M$^2$LG-DG framework. ACC, AUC, and F1 were computed separately for
Sites~9, 15, 20, and 21 and then averaged across the four target sites over
five random seeds. The results are presented in Table~\ref{tab:ablation}.

\begin{table*}[t]
\centering
\caption{Ablation study (\%) of the input modalities and the main components
of M$^2$LG-DG on the REST-meta-MDD dataset under the site-wise source-only
evaluation protocol. Results are averaged across Site~9, Site~15, Site~20,
and Site~21 over five random seeds. Bold and underlined values denote the
best and second-best results across all configurations.}
\label{tab:ablation}
\small
\renewcommand{\arraystretch}{1.12}
\setlength{\tabcolsep}{4.5pt}

\begin{tabular}{@{}llccccc|ccc@{}}
\hline
Config & Variant & Dual & Non-img & Decomp & CrossAttn &
Contrastive & ACC & AUC & F1 \\
\hline

\multicolumn{10}{c}{\textit{Input modality analysis}} \\
\hline

M1 & Non-imaging-only
& $\times$ & \checkmark & $\times$ & $\times$ & $\times$
& 53.38 & 56.70 & 52.83 \\

M2 & fMRI-only
& \checkmark & $\times$ & $\times$ & $\times$ & $\times$
& 57.85 & 61.92 & 55.74 \\

\hline
\multicolumn{10}{c}{\textit{Component analysis}} \\
\hline

C1 & w/o Decomp
& \checkmark & \checkmark & $\times$ & $\times$\textsuperscript{\dag} & \checkmark
& 59.96 & 64.21 & 57.87 \\

C2 & w/o CrossAttn
& \checkmark & \checkmark & \checkmark & $\times$\textsuperscript{*} & \checkmark
& 60.85 & 64.92 & 59.74 \\

C3 & w/o Contrastive
& \checkmark & \checkmark & \checkmark & \checkmark & $\times$
& \underline{62.31} & \underline{67.84} & \underline{61.28} \\

C4 & M$^2$LG-DG
& \checkmark & \checkmark & \checkmark & \checkmark & \checkmark
& \textbf{63.64} & \textbf{69.48} & \textbf{63.62} \\

\hline
\end{tabular}

\vspace{1mm}
\begin{minipage}{0.94\textwidth}
\footnotesize
\textsuperscript{\dag}Removing the shared-private decomposition also removes
the cross-attention fusion, since cross-attention operates on the decomposed
tokens. In this configuration, the imaging and non-imaging embeddings are
fused by concatenation followed by linear projection.\\
\textsuperscript{*}The bidirectional cross-attention module and the learned
modality gate are replaced with concatenation of the four decomposed
components followed by linear projection.
\end{minipage}
\end{table*}

\begin{itemize}

\item \textbf{Input modality analysis (M1, M2).}
The non-imaging variables alone (M1) achieved 53.38\% ACC, 56.70\% AUC, and
52.83\% F1, indicating that age, sex, and years of education provide modest
diagnostic information when considered independently. The fMRI-only configuration
(M2) achieved 57.85\% ACC, 61.92\% AUC, and 55.74\% F1. In M2, the non-imaging
encoder is removed together with the decomposition losses, cross-attention
fusion, and contrastive objective. Its performance therefore reflects the joint
absence of these components rather than the contribution of the dual-stream
encoder alone.

\item \textbf{Effect of modality access relative to the comparison methods.}
Configuration~C1 uses both modalities together with the cross-site contrastive
objective, while combining their representations through concatenation and
linear projection without the proposed decomposition and attention-based fusion.
It achieved 64.21\% AUC, compared with MSSTAN (65.88\%), M$^2$DC (66.90\%),
MetaExplainer (67.25\%), and MM-GTUNets (67.30\%) in
Table~\ref{tab:overall_rest_meta_mdd}. This observation indicates
that access to non-imaging variables alone does not explain the performance of
the complete framework. The additional 5.27 percentage points in AUC from C1 to
C4 are therefore associated with the proposed strategy for representing and
combining the two modalities.

\item \textbf{Shared-private decomposition (C1 $\rightarrow$ C2).}
Introducing $\mathcal{L}_{\text{sim}}$, $\mathcal{L}_{\text{orth}}$, and
$\mathcal{L}_{\text{diff}}$ while retaining concatenation-based fusion increased
ACC from 59.96\% to 60.85\%, AUC from 64.21\% to 64.92\%, and F1 from 57.87\%
to 59.74\%, corresponding to improvements of 0.89, 0.71, and 1.87 percentage
points, respectively. These results show that the shared-private decomposition
provides consistent improvements even when it is evaluated with the
concatenation-based fusion setting.

\item \textbf{Cross-attention fusion and modality gating (C2 $\rightarrow$ C4).}
Replacing concatenation with bidirectional cross-attention and the learned
modality gate increased ACC by 2.79, AUC by 4.56, and F1 by 3.88 percentage
points. Since C1, C2, and C4 form a nested sequence, the overall difference
between C1 and C4 can be separated into the contributions of decomposition and
fusion, with the fusion stage providing the larger increase across all three
metrics. The two components are structurally related because cross-attention
operates on the decomposed tokens. The shared-private structure therefore
provides the representations required for effective attention-based interaction
between the two modalities.

\item \textbf{Cross-site supervised contrastive loss (C3 $\rightarrow$ C4).}
Removing $\mathcal{L}_{\text{con}}$ resulted in 62.31\% ACC, 67.84\% AUC, and
61.28\% F1, corresponding to differences of 1.33, 1.64, and 2.34 percentage
points relative to C4. C3 remains the second-highest configuration across all
three metrics, showing that the multi-modal architecture provides a substantial
contribution to classification performance, while the contrastive objective
provides an additional improvement by encouraging same-class representations
from different source sites to remain close in the learned feature space.

\item \textbf{Complete framework (C4).}
M$^2$LG-DG achieved the highest values across all three evaluation metrics, with
63.64\% ACC, 69.48\% AUC, and 63.62\% F1. The complete configuration consistently
exceeded the corresponding reduced configurations, indicating that its
performance is associated with the coordinated use of both modalities,
shared-private representation learning, attention-based fusion, and the
cross-site supervised contrastive objective, rather than with the inclusion of
the additional input modality alone.

\end{itemize}

\begin{figure*}[htbp]
\centerline{\includegraphics[width=1.0\textwidth]{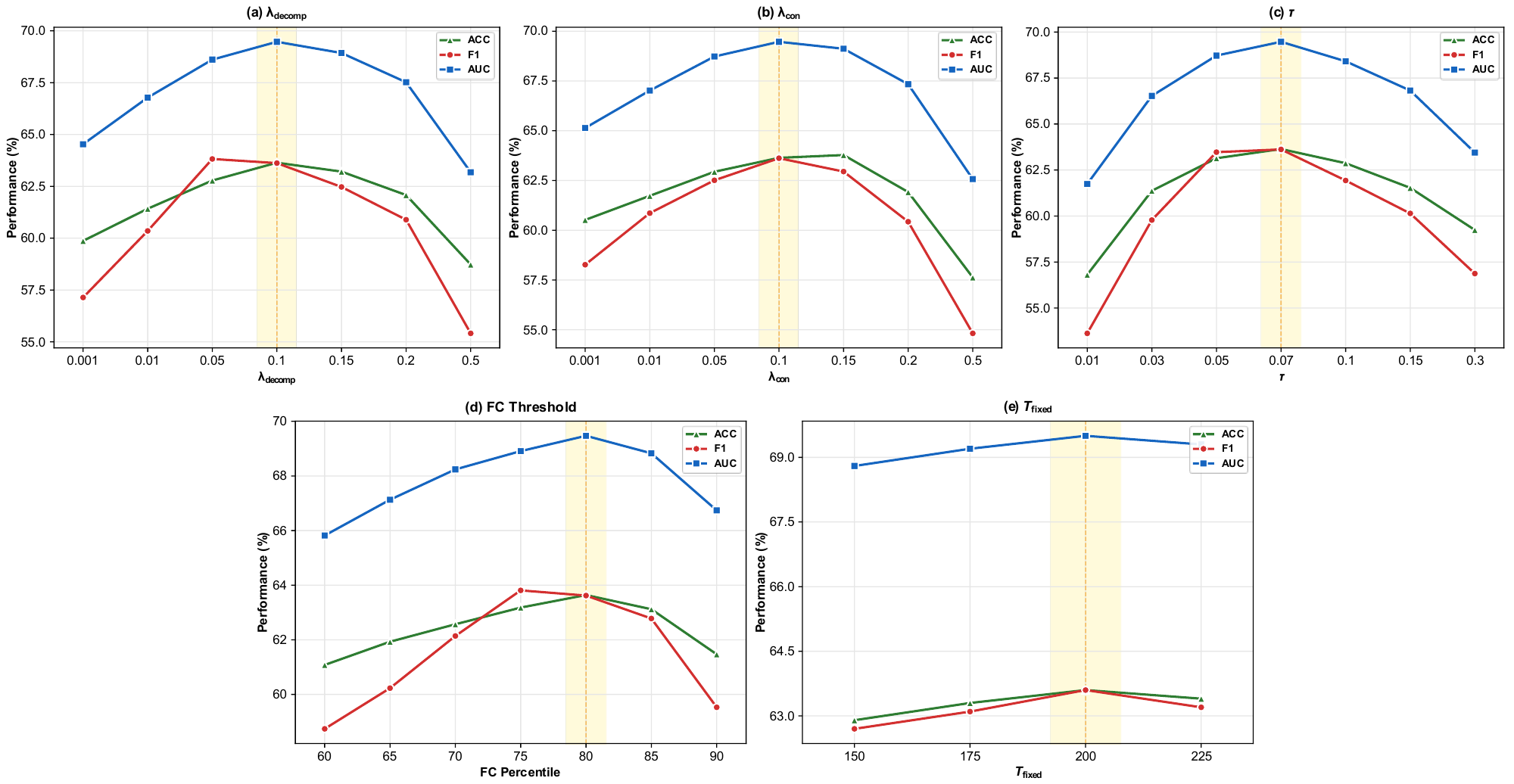}}
\caption{Sensitivity analysis of hyperparameters. The curves show the mean ACC, F1, and AUC averaged across Site~9, Site~15, Site~20, and Site~21. The yellow band indicates the selected default value.}
\label{fig:sensitivity}
\end{figure*}

\subsection{Sensitivity analysis of hyperparameters}
\label{sec:sensitivity_study}

We evaluated the sensitivity of M$^2$LG-DG to five hyperparameters: the
decomposition loss weight $\lambda_{\text{decomp}}$, the contrastive loss weight
$\lambda_{\text{con}}$, the contrastive temperature $\tau$, the FC threshold
percentile, and the temporal resampling length $T_{\mathrm{fixed}}$. The default
values were determined using source-domain validation data only and were fixed
before evaluation on any target site. Therefore, the sensitivity results were
not used for hyperparameter or model selection. In each experiment, one
hyperparameter was varied while all others were fixed at their default values,
with $\lambda_{\text{sim}} = \lambda_{\text{orth}} = \lambda_{\text{diff}} =
\lambda_{\text{decomp}}$ throughout. The results were averaged across the four
held-out target sites using the same site-wise evaluation protocol and are
presented in Fig.~\ref{fig:sensitivity}.

The two loss weights were varied over \{0.001, 0.01, 0.05, 0.1, 0.15,
0.2, 0.5\}. For $\lambda_{\text{decomp}}$
(Fig.~\ref{fig:sensitivity}(a)), performance increased up to 0.1, where
the highest ACC and AUC were observed, and then decreased gradually for
larger values. For $\lambda_{\text{con}}$
(Fig.~\ref{fig:sensitivity}(b)), the highest AUC and F1 were obtained at
0.1, while ACC remained stable between 0.05 and 0.15. These results
indicate that moderate values of both loss weights provide an appropriate
balance, whereas smaller and larger values are associated with lower
performance.

The contrastive temperature $\tau$ (Fig.~\ref{fig:sensitivity}(c)) was
evaluated over \{0.01, 0.03, 0.05, 0.07, 0.1, 0.15, 0.3\}. The highest
ACC and AUC were observed at 0.07, with lower performance toward both
ends of the evaluated range. The FC threshold percentile
(Fig.~\ref{fig:sensitivity}(d)) was varied over \{60, 65, 70, 75, 80, 85,
90\}. Performance remained stable between the 70th and 85th percentiles,
indicating limited sensitivity of the local encoder to moderate
variations in graph density.

\label{sec:temporal_length_sensitivity}
The temporal resampling length $T_{\mathrm{fixed}}$ determines the temporal
resolution provided to the global pathway. Varying $T_{\mathrm{fixed}}$ over
$\{150, 175, 200, 225\}$ (Fig.~\ref{fig:sensitivity}(e)) resulted in ACC values
between 62.9\% and 63.6\%, AUC values between 68.8\% and 69.5\%, and F1 values
between 62.7\% and 63.6\%, with the highest value of each metric obtained at
200. The maximum variation in AUC was 0.7 percentage points, which is small
relative to the between-seed variation reported in
Table~\ref{tab:internal_validation_rest_meta_mdd}. Accordingly, the vertical
scale of Fig.~\ref{fig:sensitivity}(e) differs from those of the other panels.

Based on the source-domain validation procedure, the default configuration was
set to $\lambda_{\text{decomp}} = 0.1$, $\lambda_{\text{con}} = 0.1$,
$\tau = 0.07$, the 80th percentile FC threshold, and
$T_{\mathrm{fixed}} = 200$. These settings were used in all main experiments
and ablation studies. Under this configuration, the framework achieved
approximately 63.6\% ACC, 69.5\% AUC, and 63.6\% F1, while F1 remained within
0.7 percentage points of the highest value observed across the corresponding
parameter sweeps.

\begin{figure}
    \centering
    \includegraphics[width=\columnwidth]{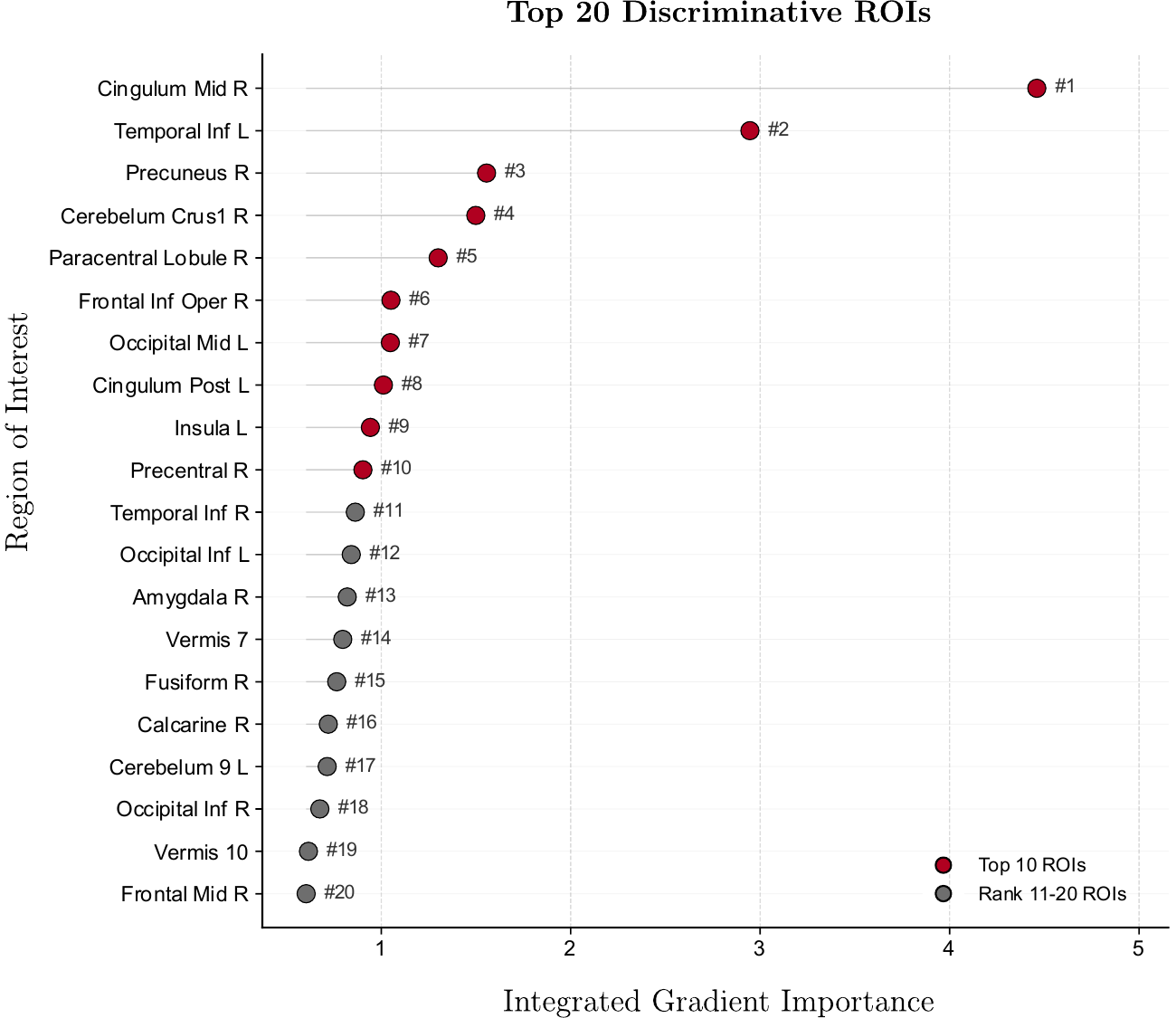}
    \caption{Top 20 ROIs identified by M$^{2}$LG-DG based on integrated gradients attribution scores. The top 10 ROIs are highlighted in red, while ranks 11 to 20 are shown in gray. Higher scores indicate a larger contribution to the model's classification decision.}
    \label{fig:brain_regions}
\end{figure}

\subsection{Analysis of important brain regions}
\label{sec:brain_regions}

To examine the brain regions contributing to MDD classification, integrated gradients were used to compute ROI-level attribution scores based on the AAL atlas~\cite{sundararajan2017axiomatic}. The attribution scores were estimated through the complete M$^{2}$LG-DG architecture, from the predicted logit through the classifier, cross-attention fusion, shared-private decomposition, and both fMRI encoders to the raw input. For each subject, the absolute attribution values were averaged across the temporal dimension to obtain an importance score for each ROI. These scores were then averaged across all test subjects and evaluation folds. The 20 ROIs with the highest average attribution scores are presented in Fig.~\ref{fig:brain_regions}, where the top 10 ROIs are highlighted in red and ranks 11 to 20 are shown in gray.

The ten highest-ranked ROIs were Cingulum\_Mid\_R, Temporal\_Inf\_L, Precuneus\_R, Cerebelum\_Crus1\_R, Paracentral\_Lobule\_R, Frontal\_Inf\_Oper\_R, Occipital\_Mid\_L, Cingulum\_Post\_L, Insula\_L, and Precentral\_R. These regions span cingulate, temporal, default-mode, cerebellar, frontal, visual, insular, and sensorimotor systems, indicating that the classification decisions were associated with distributed functional systems rather than a single anatomical region.

The identified ROIs are consistent with previous neuroimaging findings in MDD. Altered connectivity in the middle and posterior cingulate regions has been associated with affective monitoring and default-mode processing~\cite{greicius2007resting}. The right precuneus is a central component of the default-mode network, which has shown altered connectivity in patients with MDD~\cite{yan2019reduced,kaiser2015large}. Abnormal activation in the inferior temporal gyrus has also been reported during emotional processing~\cite{groenewold2013emotional}. The involvement of Cerebelum\_Crus1\_R is consistent with evidence linking posterior cerebellar regions to cognitive and affective processing in MDD~\cite{guo2015increased,liu2012altered,depping2018cerebellar}. The paracentral lobule and precentral gyrus indicate possible sensorimotor involvement~\cite{song2022abnormal}, while the left insula represents a central region of the salience network~\cite{hamilton2012functional}. The remaining highly ranked ROIs include visual, limbic, cerebellar, and frontal regions, consistent with distributed cortical alterations reported across MDD cohorts~\cite{schmaal2017cortical}. Overall, the attribution results indicate that M$^{2}$LG-DG identifies distributed brain regions that have previously been associated with MDD.

\begin{table}[t]
\centering
\caption{Classification results (\%) on the ABIDE dataset, reported as mean $\pm$ standard deviation over five random seeds. NYU and USM were each held out separately as unseen target domains, while the remaining 19 of the 20 included sites were used as source domains for model development. The best result in each column is shown in bold and the second-best is underlined.}
\label{tab:ABIDE}
\setlength{\tabcolsep}{3pt}
\renewcommand{\arraystretch}{1.1}
\resizebox{\columnwidth}{!}{%
\begin{tabular}{ll*{5}{c}}
\toprule
\multicolumn{7}{c}{\textbf{ABIDE dataset}} \\
\midrule
Site & Method & {ACC} & {SPE} & {SEN} & {AUC} & {F1} \\
\midrule

\multirow{14}{*}{NYU}
& BrainNetCNN   & 56.49 $\pm$ 3.22 & 62.24 $\pm$ 3.15 & 48.77 $\pm$ 3.95 & 63.15 $\pm$ 1.09 & 48.90 $\pm$ 3.83 \\
& CDANN         & 57.31 $\pm$ 4.24 & 62.24 $\pm$ 5.73 & 50.68 $\pm$ 3.36 & 65.35 $\pm$ 1.33 & 50.39 $\pm$ 3.89 \\
& Mixup         & 54.50 $\pm$ 2.35 & 58.78 $\pm$ 2.66 & 48.77 $\pm$ 5.00 & 65.59 $\pm$ 2.91 & 47.71 $\pm$ 3.71 \\
& MLDG          & 56.14 $\pm$ 3.33 & 58.37 $\pm$ 3.01 & 53.15 $\pm$ 4.27 & 65.49 $\pm$ 1.38 & 50.84 $\pm$ 3.85 \\
& BrainGNN      & 55.56 $\pm$ 4.22 & 61.43 $\pm$ 6.18 & 47.67 $\pm$ 3.80 & 67.89 $\pm$ 1.96 & 47.84 $\pm$ 3.79 \\
& IIB           & 58.71 $\pm$ 3.37 & 63.27 $\pm$ 5.20 & 52.60 $\pm$ 2.49 & 67.28 $\pm$ 1.46 & 52.14 $\pm$ 2.78 \\
& IBGNN+        & 58.71 $\pm$ 1.88 & 61.63 $\pm$ 3.79 & 54.79 $\pm$ 4.43 & 68.13 $\pm$ 2.60 & 53.11 $\pm$ 1.82 \\
& UFA-Net       & 57.43 $\pm$ 0.96 & 60.82 $\pm$ 3.51 & 52.88 $\pm$ 4.40 & 67.35 $\pm$ 1.47 & 51.40 $\pm$ 2.27 \\
& GenM          & 57.89 $\pm$ 2.45 & 62.86 $\pm$ 2.94 & 51.23 $\pm$ 4.71 & 67.48 $\pm$ 1.24 & 50.90 $\pm$ 3.59 \\
& MSSTAN        & 59.30 $\pm$ 2.57 & 61.63 $\pm$ 3.99 & \textbf{56.17 $\pm$ 1.38} & 68.79 $\pm$ 1.07 & \underline{54.05 $\pm$ 3.14} \\
& M$^2$DC       & 57.08 $\pm$ 2.97 & 60.20 $\pm$ 4.08 & 52.88 $\pm$ 3.15 & 67.78 $\pm$ 1.74 & 51.27 $\pm$ 2.93 \\
& MM-GTUNets    & \underline{60.23 $\pm$ 3.86} & \underline{65.51 $\pm$ 7.01} & 53.15 $\pm$ 1.79 & \underline{68.90 $\pm$ 1.38} & 53.37 $\pm$ 2.47 \\
& MetaExplainer & 59.30 $\pm$ 3.32 & 63.27 $\pm$ 3.15 & 53.97 $\pm$ 3.11 & 68.55 $\pm$ 3.10 & 53.09 $\pm$ 3.81 \\
\cmidrule(lr){2-7}
& \textbf{M$^2$LG-DG}
& \textbf{63.74 $\pm$ 4.22}
& \textbf{70.00 $\pm$ 17.53}
& \underline{55.34 $\pm$ 14.77}
& \textbf{69.20 $\pm$ 3.56}
& \textbf{56.59 $\pm$ 4.33} \\

\midrule

\multirow{14}{*}{USM}
& BrainNetCNN   & 55.74 $\pm$ 3.07 & 55.65 $\pm$ 5.67 & 55.79 $\pm$ 5.06 & 63.48 $\pm$ 3.23 & 61.01 $\pm$ 3.69 \\
& CDANN         & 58.36 $\pm$ 6.52 & 57.39 $\pm$ 3.64 & 58.95 $\pm$ 8.65 & 67.61 $\pm$ 4.23 & 63.60 $\pm$ 7.23 \\
& Mixup         & 50.82 $\pm$ 6.24 & 53.04 $\pm$ 7.78 & 49.47 $\pm$ 7.30 & 62.60 $\pm$ 1.47 & 55.48 $\pm$ 6.87 \\
& MLDG          & 54.43 $\pm$ 5.73 & 50.43 $\pm$ 2.38 & 56.84 $\pm$ 9.60 & 65.57 $\pm$ 3.45 & 60.55 $\pm$ 6.91 \\
& BrainGNN      & 51.80 $\pm$ 5.87 & 56.52 $\pm$ 6.15 & 48.95 $\pm$ 7.81 & 60.89 $\pm$ 3.35 & 55.67 $\pm$ 6.66 \\
& IIB           & 59.02 $\pm$ 5.05 & 59.13 $\pm$ 4.96 & 58.95 $\pm$ 7.35 & 67.46 $\pm$ 1.40 & 64.02 $\pm$ 5.67 \\
& IBGNN+        & 57.38 $\pm$ 4.64 & 57.39 $\pm$ 7.14 & 57.37 $\pm$ 6.81 & 69.11 $\pm$ 3.28 & 62.52 $\pm$ 5.04 \\
& UFA-Net       & 59.67 $\pm$ 6.62 & \underline{66.09 $\pm$ 3.64} & 55.79 $\pm$ 9.00 & 68.57 $\pm$ 2.14 & 63.03 $\pm$ 7.72 \\
& GenM          & 58.36 $\pm$ 9.00 & 60.87 $\pm$ 6.87 & 56.84 $\pm$ 5.13 & 70.21 $\pm$ 2.83 & 62.87 $\pm$ 2.52 \\
& MSSTAN        & \underline{61.97 $\pm$ 3.55} & \textbf{68.48 $\pm$ 1.07} & \underline{61.58 $\pm$ 3.00} & 70.36 $\pm$ 2.43 & \underline{66.87 $\pm$ 2.61} \\
& M$^2$DC       & 57.38 $\pm$ 4.78 & \textbf{68.48 $\pm$ 10.01} & 56.84 $\pm$ 5.13 & 71.19 $\pm$ 1.02 & 62.39 $\pm$ 4.24 \\
& MM-GTUNets    & 58.69 $\pm$ 3.55 & 59.13 $\pm$ 7.28 & 58.42 $\pm$ 2.88 & \underline{71.96 $\pm$ 3.65} & 63.80 $\pm$ 2.80 \\
& MetaExplainer & 57.70 $\pm$ 4.40 & 60.00 $\pm$ 8.36 & 56.32 $\pm$ 5.77 & 69.50 $\pm$ 2.54 & 62.31 $\pm$ 4.59 \\
\cmidrule(lr){2-7}
& \textbf{M$^2$LG-DG}
& \textbf{66.56 $\pm$ 3.77}
& 61.74 $\pm$ 17.23
& \textbf{69.47 $\pm$ 10.12}
& \textbf{74.14 $\pm$ 2.79}
& \textbf{72.15 $\pm$ 4.23} \\

\bottomrule
\end{tabular}%
}
\end{table}

\subsection{Experiments on model generalization}

To evaluate the applicability of M$^2$LG-DG beyond MDD classification, we conducted autism spectrum disorder (ASD) classification experiments on the multi-site ABIDE dataset \cite{di2014autism}. The analyzed cohort comprised 884 subjects from 20 international acquisition sites, including 476 HCs and 408 subjects with ASD. The rs-fMRI data were preprocessed using the C-PAC pipeline, and regional time series were extracted from 116 ROIs using the AAL atlas~\cite{tzourio2002automated}. To evaluate generalization across target domains of different sizes, NYU (171 subjects) and USM (61 subjects) were selected as held-out target domains on the basis of cohort size alone, before any model was trained on ABIDE and before any target-site performance was observed. NYU is the largest site in the analyzed cohort, whereas USM is a moderately sized site. Each site was held out separately for testing, while the remaining 19 sites were used as source domains for model development. For each experiment, M$^2$LG-DG was randomly initialized and trained from scratch using only the corresponding source-domain data, without transferring weights from REST-meta-MDD. The non-imaging features included sex, age, verbal IQ, performance IQ, and full-scale IQ. The architecture and training hyperparameters followed those used for REST-meta-MDD in Table~\ref{tbl:hyperparams}, without dataset-specific tuning.

As reported in Table~\ref{tab:ABIDE}, M$^2$LG-DG achieved the highest ACC, SPE, AUC, and F1 on NYU, with values of 63.74\%, 70.00\%, 69.20\%, and 56.59\%, respectively, together with the second-highest SEN of 55.34\%. On USM, it achieved the highest ACC, SEN, AUC, and F1 values of 66.56\%, 69.47\%, 74.14\%, and 72.15\%, respectively. Averaged across the two held-out target sites, M$^2$LG-DG obtained 65.15\% ACC, 65.87\% SPE, 62.41\% SEN, 71.67\% AUC, and 64.37\% F1, supporting its applicability to other multi-site neuroimaging classification tasks.

\section{Discussion}
\label{sec:discussion}

In this study, M$^2$LG-DG addresses cross-site MDD classification by learning
diagnostic representations that remain effective under variations in acquisition
protocols, scanner characteristics, and cohort composition across imaging
centers. As reported in Table~\ref{tab:internal_validation_rest_meta_mdd}, the
framework achieves the highest AUC at all four held-out sites, with values of
75.81\%, 68.40\%, 70.16\%, and 63.53\%, respectively, and an average AUC of
69.48\%. AUC provides a useful measure in this setting because it does not
depend on a particular decision threshold. The threshold-dependent metrics
exhibit greater variation across sites. Site~9 shows higher specificity
(85.20\% SPE, 46.40\% SEN), whereas Site~21 shows higher sensitivity
(75.12\% SEN, 37.71\% SPE). A similar pattern is observed for the comparison
methods at these sites, suggesting that the differences in sensitivity and
specificity are related to variations in class composition and acquisition
conditions among the target cohorts rather than to the behavior of a single
model.

The ablation study in Table~\ref{tab:ablation} further illustrates the
contribution of the main components of M$^2$LG-DG. Introducing the
shared-private decomposition losses with concatenation-based fusion
(C1 $\rightarrow$ C2) improves ACC, AUC, and F1 by 0.89, 0.71, and 1.87
percentage points, respectively. Replacing concatenation with bidirectional
cross-attention and modality gating (C2 $\rightarrow$ C4) provides additional
improvements of 2.79, 4.56, and 3.88 percentage points. Since C1, C2, and C4
form a nested sequence, the overall improvements of 3.68, 5.27, and 5.75
percentage points can be separated into the contributions of decomposition and
fusion, with the fusion stage accounting for the larger improvement across the
three metrics. These components are also structurally related because the
cross-attention module operates on the decomposed representations. The
shared-private decomposition therefore provides structured modality
representations that are subsequently integrated through attention-based
fusion. When the cross-site supervised contrastive loss is omitted, ACC, AUC,
and F1 change from 63.64\%, 69.48\%, and 63.62\% to 62.31\%, 67.84\%, and
61.28\%, respectively, corresponding to differences of 1.33, 1.64, and 2.34
percentage points.

The input modality analysis provides additional information regarding the role
of non-imaging variables. The non-imaging-only configuration M1 achieves
56.70\% AUC, whereas configuration C1, which combines both modalities using
the contrastive objective and concatenation-based fusion, achieves 64.21\%
AUC. This value remains below those of the four leading comparison methods in
Table~\ref{tab:overall_rest_meta_mdd}. These observations suggest that the use
of age, sex, and years of education alone does not account for the overall
performance of M$^2$LG-DG. This finding is also consistent with the comparison
with MM-GTUNets~\cite{cai2025mm}, which similarly incorporates non-imaging
variables. The sensitivity analysis in Fig.~\ref{fig:sensitivity} shows stable
performance for moderate values of $\lambda_{\text{con}}$ and $\tau$, for FC
thresholds between the 70th and 85th percentiles, and across the evaluated
values of $T_{\text{fixed}}$. In particular, the AUC varies by no more than
0.7 percentage points across the examined temporal resampling lengths.
Furthermore, Fig.~\ref{fig:brain_regions} identifies distributed regions
associated with cingulate, default-mode, cerebellar, sensorimotor, and salience
systems. The additional experiments on ABIDE, with average results of 65.15\%
ACC, 71.67\% AUC, and 64.37\% F1, also indicate that the proposed framework
can be applied to another multi-site neuroimaging classification setting.

In future studies, we plan to investigate acquisition-aware temporal standardization to account more explicitly for differences in repetition time, scan duration, and temporal sampling across sites. This direction could also be integrated with federated learning, where temporal preprocessing and model
training are performed locally at participating centers while model parameters
are shared for collaborative optimization. Such a setting would enable multi-center model development without requiring the transfer of subject-level imaging data and may provide a practical approach for extending cross-site neuroimaging analysis to distributed clinical environments.

\section{Conclusion}
This study proposes M$^2$LG-DG, a multi-modal local-global domain generalization framework for cross-site MDD classification from rs-fMRI and non-imaging features. It addresses three key challenges in multi-site diagnostic classification: incomplete utilization of complementary temporal and spatial brain activity patterns, limited integration of non-imaging information, and cross-site distribution shifts caused by scanner heterogeneity. Accordingly, the framework integrates dual-stream fMRI encoding, shared-private decomposition with auxiliary losses, bidirectional cross-attention fusion with a learned modality gate, and cross-site supervised contrastive learning for site-invariant representations. Experimental results on four held-out REST-meta-MDD sites and two held-out ABIDE sites, obtained under a source-only protocol where each target site is excluded from training, validation, and model selection, demonstrate the effectiveness of the proposed framework across the evaluated target domains.

\printcredits


\bibliographystyle{model1-num-names}

\bibliography{cas-refs}
\end{document}